\documentclass[preprint,10pt,authoryear]{elsarticle}

\usepackage{amssymb}

\usepackage{array, amsmath}
\usepackage{algorithm}
\usepackage[noend]{algorithmic}
\usepackage{xcolor}
\usepackage{multirow}
\usepackage{diagbox}
\usepackage[flushleft]{threeparttable}
\usepackage{enumitem}
\usepackage[caption=false,font=footnotesize]{subfig}

\journal{Expert Systems with Applications}

\begin{document}

\begin{frontmatter}



\title{Financial Assets Dependency Prediction Utilizing Spatiotemporal Patterns}

\author[label1]{Haoren Zhu}
\ead{hzhual@cse.ust.hk}
\author[label2]{Pengfei Zhao\corref{cor1}}
\ead{ericpfzhao@uic.edu.cn}
\author[label1]{Wilfred Siu Hung NG}
\ead{wilfred@cse.ust.hk}
\author[label1]{Dik Lun Lee}
\ead{dlee@cse.ust.hk}

\cortext[cor1]{Corresponding author}
\affiliation[label1]{organization={Hong Kong University of Science and Technology},
            country={Hong Kong}}

\affiliation[label2]{organization={BNU-HKBU United International College},
            city={Zhuhai},
            state={Guangdong},
            country={China}}



\begin{abstract}
Financial assets exhibit complex dependency structures, which are crucial for investors to create diversified portfolios to mitigate risk in volatile financial markets. To explore the financial asset dependencies dynamics, we propose a novel approach that models the dependencies of assets as an Asset Dependency Matrix (ADM) and treats the ADM sequences as image sequences. This allows us to leverage deep learning-based video prediction methods to capture the spatiotemporal dependencies among assets. However, unlike images where neighboring pixels exhibit explicit spatiotemporal dependencies due to the natural continuity of object movements, assets in ADM do not have a natural order. This poses challenges to organizing the relational assets to reveal better the spatiotemporal dependencies among neighboring assets for ADM forecasting. To tackle the challenges, we propose the Asset Dependency Neural Network (ADNN), which employs the Convolutional Long Short-Term Memory (ConvLSTM) network, a highly successful method for video prediction. ADNN can employ static and dynamic transformation functions to optimize the representations of the ADM. Through extensive experiments, we demonstrate that our proposed framework consistently outperforms the baselines in the ADM prediction and downstream application tasks. This research contributes to understanding and predicting asset dependencies, offering valuable insights for financial market participants.
\end{abstract}

\begin{graphicalabstract}
\end{graphicalabstract}

\begin{highlights}
\item Model asset correlations as an Asset Dependency Matrix (ADM).

\item Employ static and dynamic transformation to optimize the ADM representations.

\item Leverage video prediction method to capture the spatiotemporal dependencies among assets.

\item Superior forecasting performance in ADM prediction and downstream application tasks.

\end{highlights}


\end{frontmatter}


\section{Introduction}\label{sec:intro}
Financial assets, such as stocks, securities, and derivatives, exhibit diverse interdependencies, giving rise to a complex structure of dependencies among these assets \citep{elton1973estimating,ane2003dependence}. 
For instance, stocks within the same industry often exhibit a collective response to market news, causing them to move in a synchronized manner.
Moreover, the price fluctuations observed in one industry at a specific time can subsequently impact its downstream industries.
The dependency structure of the financial market is very complex, and the ability to predict it results in significant financial advantages. 
As an illustration, one practical approach to mitigate portfolio risk is through diversification across distinct asset classes that exhibit either independence or negative correlation. This strategy safeguards against the simultaneous loss of all assets in the event of a negative occurrence, such as an unfavorable news release.

The measurement of asset dependency encompasses various methodologies, such as analyzing the correlation and covariance coefficients between two assets.
When considering a collection of assets, these pairwise dependencies can be systematically arranged within a matrix, termed \textit{Asset Dependency Matrix} (\textit{ADM}). For instance, when employing the correlation coefficient as a measure of asset dependency, the {ADM} corresponds to the widely recognized correlation matrix.
Various statistical methods have been proposed for predicting future {ADM}s. 
A simple prediction model \citep{elton1978betas} uses the correlation matrix computed from the time window ending at $t$ as the predicted {ADM} for $t' > t$.  
Subsequent advancements have proposed enhancements considering asset groups, financial indexes, and beta estimations \citep{ elton1977simple,baesel1974assessment,blume1975betas,vasicek1973note}. 
A recent study presented a methodology utilizing the Cholesky decomposition and support vector regression for dynamic modeling and forecasting of covariance matrices \citep{bucci2022comparing}. 
This methodology ensures the positive definiteness of forecasted covariance matrices, which are leveraged by different parametrizations and methodologies for accurate covariance matrix forecasting \citep{verzelen2010adaptive, damodaran1999estimating}.
However, prevailing financial solutions suffer from restrictive statistical assumptions and oversimplify time-varying dependence structures to linear or approximate linear relationships between variables. Furthermore, these solutions often encounter computational inefficiencies when dealing with large matrices, thereby impeding their scalability. Thus, deep learning models that are good at modeling nonlinear relationships and time-dependent patterns are a natural fit.

Asset dependencies exhibit both temporal and relational properties. 
Specifically, fluctuations in the return of one asset, triggered by specific events, have a consequential impact on the future price movements of both itself and other assets, thus indicating {temporal dependency}.
In contrast, relational dependency pertains to the intricate interconnections among a set of assets, causing them to respond similarly to financial events.
For example, the stocks in the high-tech sector are inherently correlated (similar) because of the common growth and high-leverage nature of the companies which tend to respond similarly to events such as an increase in interest rates. The evolution of the financial market over time is influenced by both temporal and relational dependencies among assets. A notable illustration is the phenomenon known as ``sector rotation'' in finance, where investors strategically adjust their investments across various sectors of the economy to leverage the diverse performance of sectors during different phases of the economic cycle. This strategic approach allows investors to capitalize on the shifting dynamics of sectors, aligning their portfolios with the prevailing economic conditions.

\begin{figure}
  \centering
  \subfloat[Assets are randomly posited.]{\includegraphics[width=0.8\linewidth]{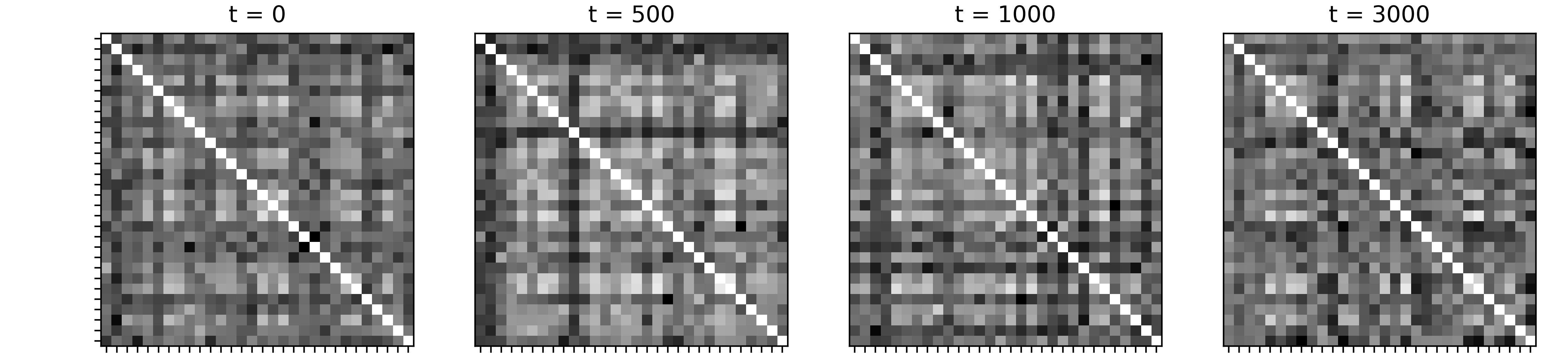}} \\
  \subfloat[Same industry assets are posited together.]{\includegraphics[width=0.8\linewidth]{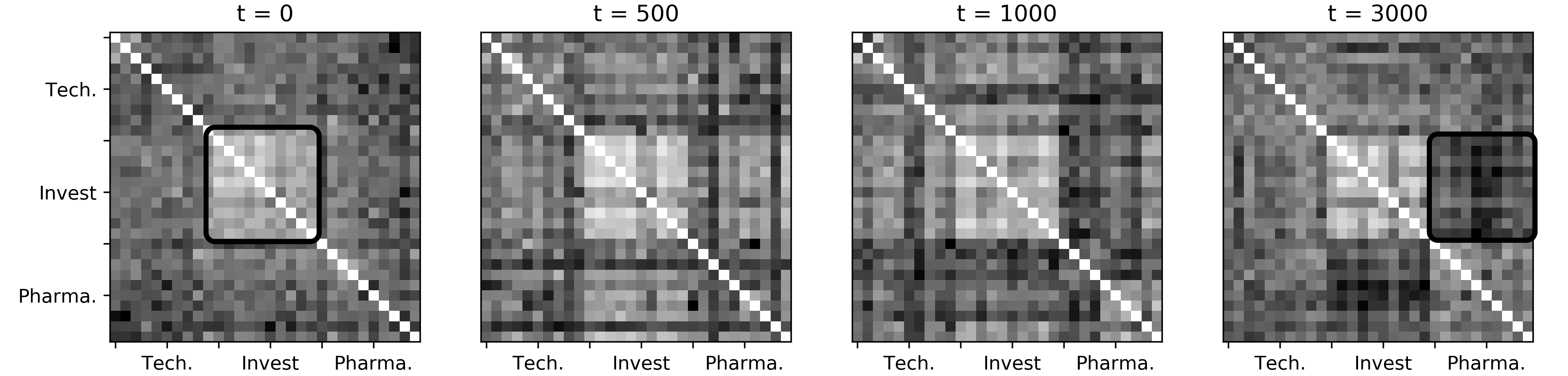}}
  
  \caption{Correlation matrix of multiple assets: (a) assets are randomly posited and (b) assets from the same industry are posited together.}
  \label{fig:intro}
\end{figure}

In this paper, we formulate the sequential {ADM} prediction task as forecasting the future {ADM} based on a sequence of past {ADM}s. We propose solutions inspired by video prediction problems \citep{Oprea_2020}, aiming at predicting future image frames based on a sequence of past frames. Each frame corresponds to an {ADM} matrix, with each frame pixel representing the correlation between two assets. Video frame data are characterized by spatiotemporal properties since a pixel's value is related to its neighboring pixels and pixels in previous frames. Deep learning-based spatiotemporal models, e.g., Convolutional Recurrent Neural Network (\textit{CRNN}), have been successful due to their ability to extract features in both spatial and temporal dimensions \citep{shi2015convolutional,mathieu2015deep,hsieh2018learning,babaeizadeh2017stochastic,wang2018eidetic}. A major challenge of directly applying {CRNN} to {ADM} prediction is that \textit{pixels of the same object are naturally grouped spatially in the images, while the correlation coefficients between the assets do not have predefined placement in the ADM}. 
In Figure \ref{fig:intro}(a), assets are placed randomly in the matrix. We can see there are no clear asset dependencies shift patterns across the sequence of frames. Thus, {CRNN} will not be effective if it directly takes {ADM}s with arbitrarily placed assets as input. Hence, it is imperative to devise suitable transformation strategies to optimally align the assets for the disclosure of the evolution pattern of {ADM}. This is essential for enhancing the accuracy of forecasting future {ADM}. Since the transformation explicitly or implicitly repositions the assets in the {ADM}, we term the locality patterns within the transformed ADM as assets' \textit{spatial dependencies}.

Figure \ref{fig:intro}(b) illustrates the idea. Ten assets are selected from each of the technology, investment, and pharmaceutical sectors. Assets from the same industry are placed next to each other in the matrix (i.e., having sequential indexes). Each matrix entry represents the correlation coefficient between two assets, where white indicates a coefficient of 1 and black a coefficient of -1. At $t=0$, we can observe that the center region of the {ADM}, representing dependencies between assets in the ``invest'' sector, has a light color and hence positive correlations. On the other hand, the region to the right of the center, representing dependencies between assets in the ``invest'' sector and assets in the ``pharmaceuticals'' sector, has a dark color, showing weakly negative correlations between the two categories of assets. Both examples illustrate various strengths of spatial dependencies between the assets. As time elapses, groups of assets show different correlations. At $t=3000$, we can see that assets in ``invest'' remain generally positively correlated, while assets in ``invest'' and assets in ``pharmaceuticals'' have become more negatively correlated (color has become darker). 
Comparing Figure \ref{fig:intro}(a) and \ref{fig:intro}(b), we can see that the proper positioning of the assets can reveal the shift of asset dependencies across time. However, re-positioning assets by grouping them in the same sector, as illustrated above, is an intuitive but ad hoc solution, since there could be many driving factors that cannot be captured by a static and explicit rule (e.g. group by industry). The exploration of spatial dependency is to find an optimal way to arrange the spatially dependent assets in the asset space so that the {CRNN} works effectively. 

In this paper, we propose the \textit{Asset Dependency Neural Network} (\textit{ADNN}), an end-to-end neural network framework that unifies the {ADM} transformation and prediction processes. 
Drawing inspiration from representation learning \citep{bengio2014representation}, we designate the task of mapping the {ADM} to its optimal representation as the \textit{ADM representation problem}. We design static and dynamic transformation paradigms to address the ADM representation problem, which can be employed individually or jointly within the {ADNN}. For the static paradigm, we design the \textit{Positional Rearrangement} algorithm that enhances {ADM}s' locality property by allocating assets within the same cluster adjacent to each other. For the dynamic paradigm, we design the \textit{MoE Transformation} block to address the time-varying relationships among multiple assets by dynamically reconfiguring the matrix arrangement over time.  
Specifically, we incorporate the \textit{Mixture of Experts} (\textit{MoE}) technique to capture the dynamic nature of market regimes, where each expert within the {MoE} framework is dedicated to modeling a specific economic period.
Finally, the transformed matrix, containing fine-grained spatiotemporal patterns, is fed into the convolution recurrent neural network blocks to generate accurate forecasting. 
Experimental results demonstrate that the {ADNN} incorporating the two proposed transformation functions outperforms baseline methods in terms of {ADM} prediction accuracy and portfolio risk reduction.

Preliminary results of this research have been published in \citep{zhu2022forecasting}. 
This comprehensive version included several new materials.
Firstly, we present the \textit{Positional Rearrangement} that solely alters the entry positions within the {ADM}, which can be used separately or combined with {MoE} transformation to achieve better prediction accuracy.
This new approach leverages consistent grouping patterns to preprocess ADMs and excels in efficiency and effectiveness when applied to well-structured data. Furthermore, it enhances interpretability by providing insights into the repositioning of matrix entries.
Secondly, we add the positive semi-definite (\textit{PSD}) enforcement block to ensure the positive semi-definiteness property of the predicted {ADM}s, which is a common requirement in the finance domain. 
Thirdly, for the experiment, we add extensive ablation studies to validate the efficacy of the proposed framework components. 
We demonstrate how the mixture of experts, combined with a quadratic function in the current design, can effectively capture information from different regimes by visualizing the activation patterns of the experts in {MoE}. 
We study three market scenarios, starting from a simple one with only one market factor at each market phase to the real market data. The results conform to our expectation that {MoE} will activate a small set of experts when the market is simple and deploy more experts as the complexity of the market increases. 
Fourthly, to help understand the transformation, we visualize the {ADM}s before and after transformation.  
Finally, in addition to portfolio optimization presented in \citep{zhu2022forecasting}, we study the application of {ADNN} to pair trading.
To conclude, prior studies have primarily focused on the forecasting aspect of the {ADM}, considering it as a singular component of the overall framework. 
In contrast, this work presents a comprehensive pipeline that encompasses the entire process starting from data preprocessing to application, and substantiates the proposed framework with a broader range of experimental results.

\section{Related Work}\label{sec:related}
\subsection{Conventional ADMs Prediction Models}
Multivariate GARCH (\textit{MGARCH}) models extend univariate volatility models to estimate time-varying asset dependencies. Initial MGARCH models \citep{bollerslev1988capital, engle1995multivariate} parameterize the conditional covariance matrix as a function of past information, i.e. the lagged conditional covariance matrix and returns. Although this generalization is flexible to dynamic structures, it is seldom applied to model large-scale assets due to the high number of unknown parameters to be estimated. 
Two modified {MGARCH} models were proposed to reduce the complexity of {MGARCH} models. One is the factor {MGARCH} models \citep{engle1990asset, ng1992multi}, which assume the dependency structure of assets generated from a few common factors, thus significantly reducing the number of parameters to be estimated. However, this modification has limitations in finding factors that can explain the dependency of assets adequately and impose an infeasible assumption that the volatilities of different assets are driven by the same factors. Another modification of {MGARCH} models is the Dynamic Conditional Correlation (\textit{DCC}) model \citep{engle2002dynamic}, which allows for different dynamics of individual assets and a linear dynamic structure of the correlation matrix. {DCC} models are easy to estimate because they assume a parsimonious dynamic structure of the conditional correlation matrix, with only two parameters to describe the time-varying pattern of correlations. One drawback of this simplification is that it is unjustifiable to assume all correlations evolve similarly regardless of the involved assets \citep{bauwens2006multivariate}. In summary, {MGARCH} models balance between model flexibility and complexity by assuming specific structures on {ADM}s. Although this reduces model complexity and speeds up the estimation process, it suffers from misspecification of the dynamics of asset dependencies. Compared with existing works, our framework is more flexible since it does not require imposing structures on {ADM}s, making it capable of effectively capturing different temporal and spatial patterns.

\subsection{Convolutional Recurrent Neural Network}
\label{subsec:temp_spat_model}
Recent advances in deep learning provide abundant insights into modeling spatial and temporal signals from data. For modeling the sequential temporal information, early literature often applies Recurrent Neural Network \citep{ng2015short,2014googlearticle}. Long Short Term Memory (\textit{LSTM}) \citep{HochSchm97} was then proposed to tackle the vanishing and exploding gradient problems that occurred in long sequences. For modeling spatial information from data, convolutional neural network approaches \citep{Goodfellow-et-al-2016} were proposed where spatial signals are compressed and extracted through the slide of the convolution kernel over images.
The encoding of the intertwined temporal and spatial relation is also discussed recently. Several works \citep{tran2015learning,8594916,8814249} use 3D convolutional neural networks to extract spatial and temporal features. Although the additional dimensions enable neural networks to model both spatial and temporal correlations in the data, the compressed features are still local features without considering sequential patterns. To better model spatiotemporal sequence, \textit{Convolutional LSTM} (\textit{ConvLSTM}) \citep{shi2015convolutional} was proposed to model the spatial and temporal structures simultaneously by changing all the input data into tensors, through which the limitation of modeling the spatial information in {LSTM} and the constraint of modeling global sequential information in the 3D convolutional neural network is overcome. Based on the pioneering work of {ConvLSTM}, other advanced spatiotemporal modeling techniques are also proposed. For example, \textit{PredRNN} \citep{wang2017predrnn} enables memory states to zigzag in two directions to break the constraint of memory storage in each cell unit, and \textit{Eidetic 3D LSTM} (\textit{E3D-LSTM}) \citep{wang2018eidetic} integrates 3D convolutions into its model to better store the short-term features and a gate-controlled self-attention module is designed to better model long-term relations. Due to the {ADM} representation problem introduced in Section \ref{sec:intro}, {ConvLSTM} based spatiotemporal techniques cannot be directly applied here and the transformation of {ADM}s is needed. In the AAAI version of this manuscript \citep{zhu2022forecasting}, we tackle the problem with {ADNN} which unifies the transformation and prediction blocks and enables the transformation and prediction functions to be learned simultaneously.

\subsection{Deep Learning for Portfolio Optimization}
Deep learning models for financial portfolio optimization leverage the power of neural networks to learn complex patterns and relationships in financial data. 
Numerous research \citep{jiang2017deep,ye2020reinforcement,wang2021deeptrader,hambly2023recent} have approached portfolio optimization as a strategic problem and have employed reinforcement techniques to optimize the selection of actions under different settings and constraints.
Moreover, various studies have utilized deep-learning architectures to forecast stock prices and subsequently incorporated these predictions into the portfolio optimization process. These approaches encompass classical machine learning models \citep{ma2021portfolio, chen2021mean} for direct price prediction, \textit{Graph Convolutional Network} (\textit{GCN}) models \citep{chen2021novel, ye2021multi, cheng2022financial} for relational property extraction, and hybrid models \citep{hou2021st, wu2023graph,wu2022s_i_lstm} that consider additional factors.
Recent studies \citep{imajo2021deep, uysal2023end} have adopted an end-to-end approach that learns the optimal portfolio weights directly from raw price data as the input.
\cite{zheng2023deep} forecasts dynamic correlation matrix by incorporating normalizing flows to obtain high-quality hash representations and facilitates regularizers in imposing structural constraints.
In this study, our focus lies in leveraging the spatiotemporal information embedded in the ADMs to forecast the asset dependencies within a group of assets, which is then employed for portfolio optimization.

\section{Problem Definition}\label{sec:preli}
\subsection{ADM Definition}
\label{subsec:adm_def}
In financial convention, given the historical daily prices of an asset $\{p_0, p_1, \allowbreak ..., p_T\}$, where $p_t$ denotes the price at time $t$, the log return of the asset at time $t$ is calculated by:
\begin{equation}
    r_t = log(p_t) - log(p_{t-1})
\end{equation}
where logarithm normalization is applied to the price data so that the derived returns are weakly stationary. Denote the return of asset $a_i$ at time $t$ as $r_t^{a_i}$ and $f(a_i,a_j)_t$ denote the dependency measurement between asset $a_i$ and asset $a_j$ at time $t$. 
If the sample covariance is picked, $f(a_i,a_j)_t$ in period $(t-n_{lag}, t]$ is calculated as follows:
\begin{equation}
    f(a_i,a_j)_t = \frac{1}{n_{lag} - 1}\sum_{s = 1}^{n_{lag}} (r_{t-s+1}^{a_j} - \bar{r}^{a_j})(r_{t-s + 1}^{a_i} - \bar{r}^{a_i})
\end{equation}
where $n_{lag}$ denotes the time span of calculating the asset dependency coefficients and  $\bar{r}^{a_i}$ means the average return of asset $i$ in the time interval. 
It is natural to use matrices to represent all pairwise dependencies. Hence, {ADM} at time $t$, denoted as $\mathcal{M}_t$, can be formulated as:
\begin{equation}
\setlength{\abovedisplayskip}{3pt}
\setlength{\belowdisplayskip}{3pt}
  \mathcal{M}_t =  \left[
\begin{matrix}
 f(a_i,a_i)_t      & f(a_i,a_j)_t      & \cdots & f(a_i,a_n)_t  \\
 f(a_j,a_i)_t      & f(a_j,a_j)_t      & \cdots & f(a_j,a_n)_t      \\
 \vdots & \vdots & \ddots & \vdots \\
 f(a_n,a_i)_t      & f(a_n,a_j)_t     & \cdots & f(a_n,a_n)_t    \\
\end{matrix}
\right]
\end{equation}

\subsection{ADM Prediction Model}
\label{subsec:adm_model}
\begin{figure}
  \centering
  \includegraphics[width=0.6\textwidth]{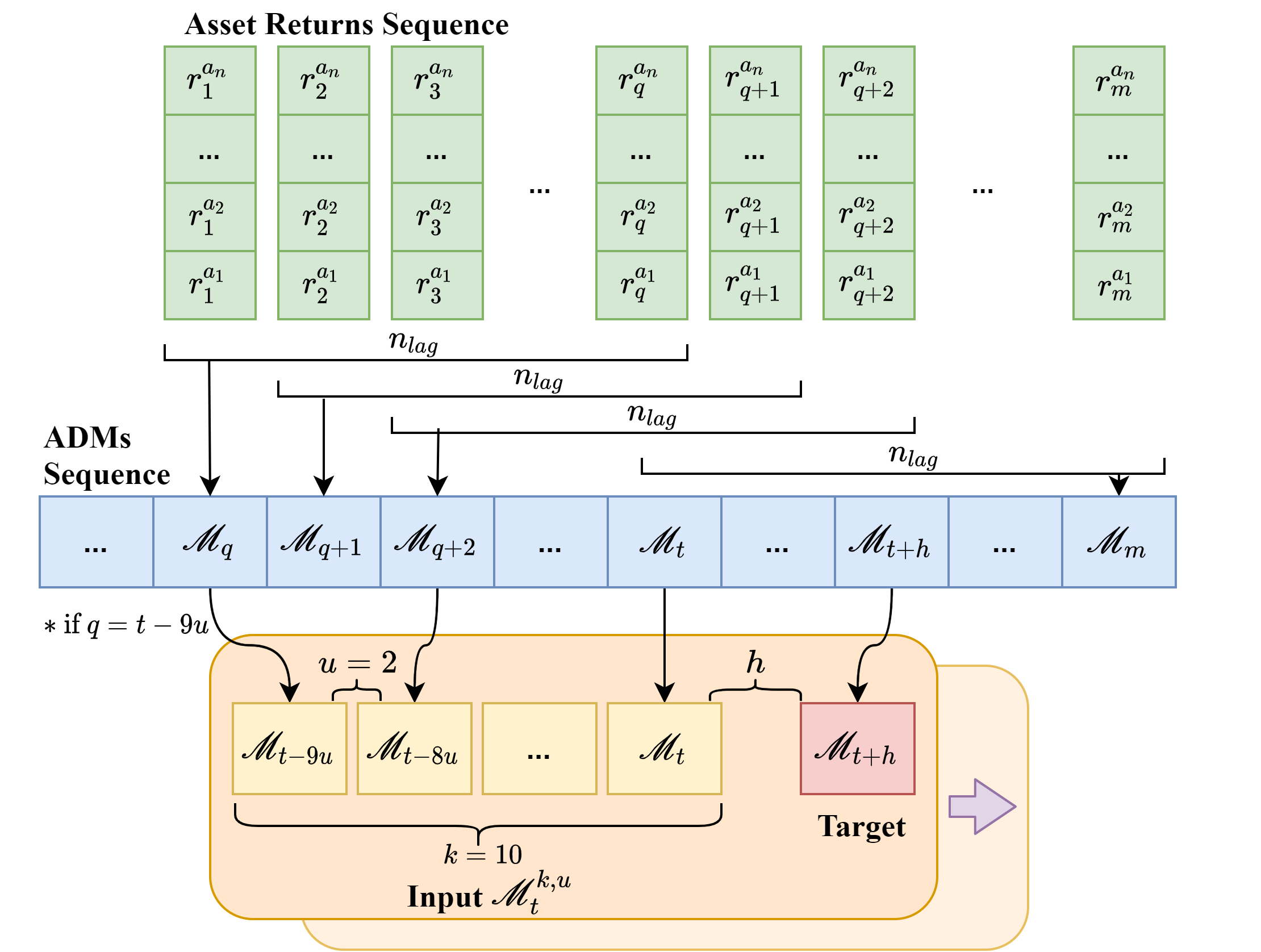} 
  \caption{{ADM}s Construction. $r_q^{a_j}$ denotes the return of asset $j$ at time $q$, and $\mathcal{M}_t$ denotes the {ADM} at time $t$. We first generate {ADM}s sequence from the asset returns sequence. In this figure, $k=10, u=2$ and $q=t-(k-1)u=t-18$. With fixed $u$, $k$, $h$, and shift of current time $t$, we can obtain multiple input sequences and their corresponding targets.} 
\label{fig:adm_structure} 
\end{figure}

Given a historical {ADM} sequence denoted as: 
\begin{equation}
\begin{aligned}
\mathcal{M}_t^{k,u} = ~\{\mathcal{M}_{t-(k-d)\cdot u} \vert 1 \leq d \leq k, t \geq (k-d) \cdot u,  \mathrm{ with }~ t,k,u,d \in Z^{+} \}
\end{aligned}
\label{equ:mtku}
\end{equation}
where $k$ denotes the length of the input sequence, $t$ denotes the current timestamp and $u$ denotes the rolling distance between each input matrix in the day unit. Figure \ref{fig:adm_structure} illustrates the construction process of the  {ADM} sequence. $\mathcal{M}_t^{k,u} \in R^{k \times c \times w \times h}$, where $c$ denotes the number of channels, $w$ and $h$, respectively, denote the width and height of {ADM}. Our target is to predict $\mathcal{M}_{t+h} \in R^{1 \times c \times w \times h}$, where $h$ denotes the horizon, which is the duration that an investor expects to hold a portfolio. 
For example, if the portfolio is rebalanced monthly, i.e., the investors are interested in the {ADM} one month later, $h$ should be set to $21$ (monthly trading days).
We formulate the {ADM} prediction model as :
\begin{equation}
    \hat{\mathcal{M}}_{t+h} = \mathcal{F} ( \mathcal{M}_t^{k,u} \vert \Theta_{\mathcal{F}} ) 
\end{equation}
where $\Theta_{\mathcal{F}}$ denotes the model parameters of $\mathcal{F}$ and $\hat{\mathcal{M}}_{t+h}$ denotes the predicted {ADM}.

\section{Methodology}\label{sec:algo}
\begin{figure}[tb]
  \centering
  \includegraphics[width=0.95\textwidth]{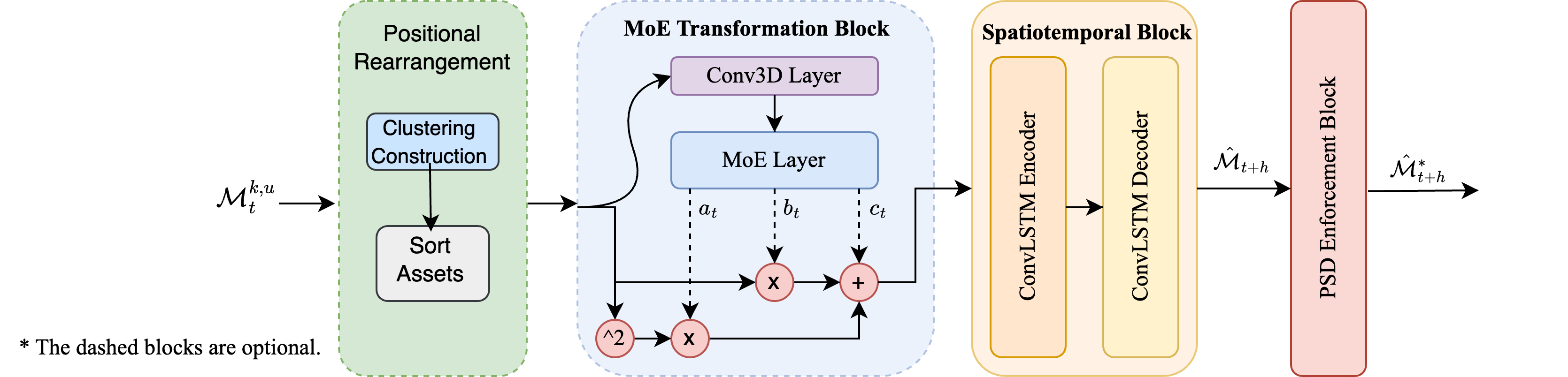} 
  \caption{The overall architecture of the {ADNN} framework. The input $\mathcal{M}_t^{k,u}$ is processed through the positional rearrangement block and the transformation block using the quadratic transformation learned by the {MoE} network. The resulting converted matrix is then passed to the spatiotemporal block to generate the forecast matrix, which is further adjusted by the PSD enforcement block to ensure positive semi-definiteness. } 
  \label{fig:general_framework} 
\end{figure}

The {ADM} forecasting problem involves two main challenges: (1) addressing the {ADM} representation problem discussed in Section \ref{sec:intro}, (2) accurately modeling the temporal and spatial signals encoded in historical {ADM}s. For the first challenge, we propose two alternative ADM transformation strategies: (1) the positional rearrangement block which statically alters the asset order in {ADM} construction process, and (2) the end-to-end transformation block which dynamically learns the optimal representations of the input {ADM}s. To address the second challenge, we propose to integrate a spatiotemporal block into the prediction task.
Figure \ref{fig:general_framework} illustrates the overall architecture. 
Note that two transformation strategies can be combined to further strengthen the ability of the prediction model. 
The details of the framework are explained in the following sections.

\subsection{Positional Rearrangement}\label{subsec:position}
As introduced in Section \ref{sec:intro}, the direct application of spatiotemporal prediction models on {ADM} sequence suffers from the {ADM} representation problem due to entries in {ADM} not having predefined placement to facilitate spatiotemporal modeling, whereas pixels in images are naturally grouped according to the  ``shape'', ``color'' and ``texture'' features \cite{ping2013review}. 
This lack of proper organization of asset dependencies obscures the information compression and boundary identification of {ADM}, making the convolution techniques less effective.
Thus, it is necessary to transform the {ADM}s so that the assets' relational features can be better revealed before the downstream convolution kernel. 
Given the established stock market sector analysis, assets within the same industry exhibit concurrent behavior. 
Our approach aims to enhance the locality property of the {ADM} by strategically rearranging the matrix columns and rows in the preprocessing stage. This positional rearrangement technique is employed to exploit the inherent relationships among assets, e.g., those belonging to the same industry or displaying concurrent patterns, further improving the effectiveness and accuracy of the {ADM} representation.

\begin{figure}[tb]
  \centering
  \setlength{\abovedisplayskip}{0pt}
  \setlength{\belowdisplayskip}{-10pt}
  \includegraphics[width=0.8\linewidth]{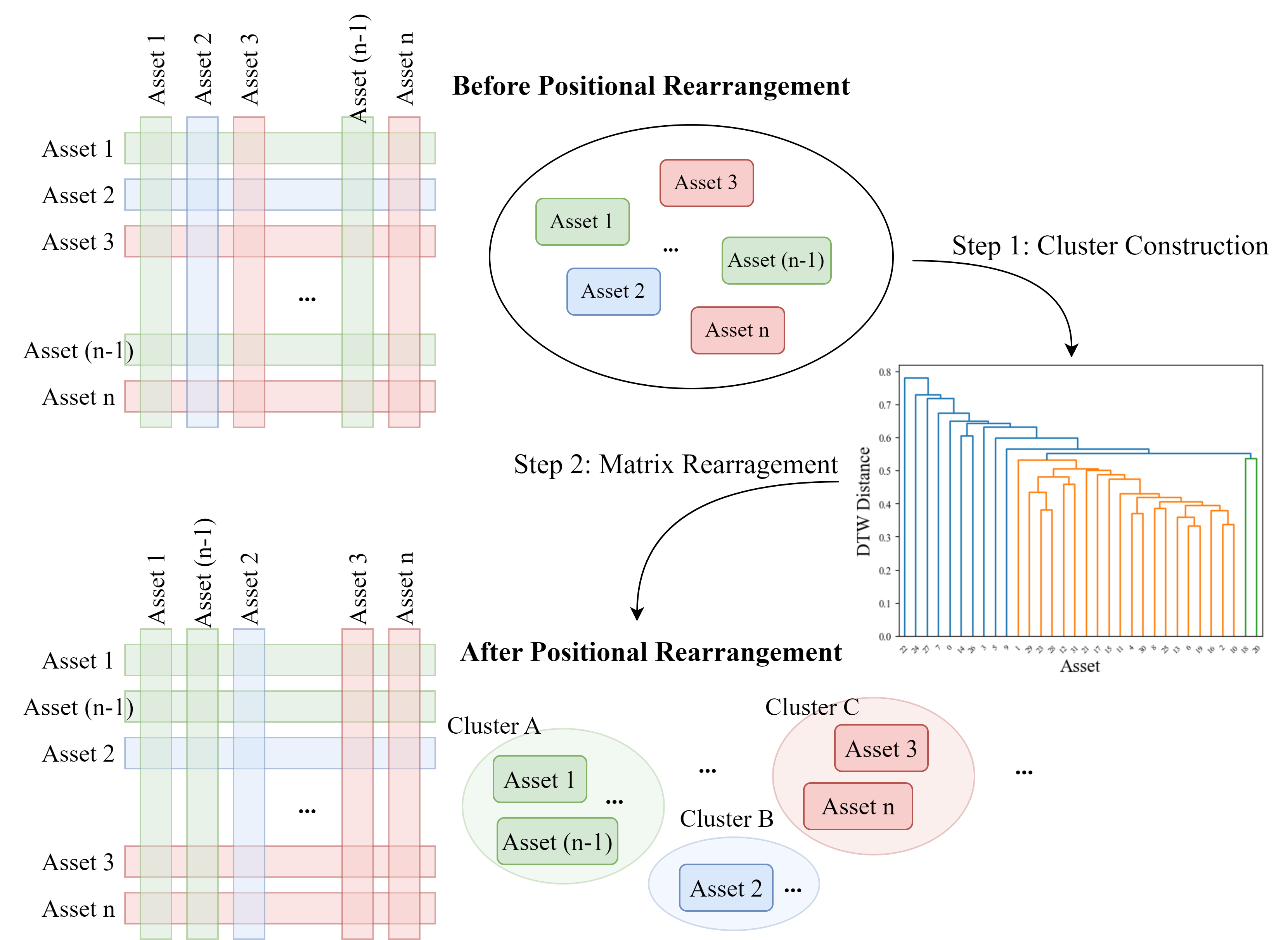} 
  \caption{Illustration of {ADM} positional rearrangement process. Before the positional rearrangement, assets inside the {ADM} are randomly positioned, resulting in poor spatial patterns. In contrast, the {ADM} after the positional rearrangement is structured based on the cluster information, thereby enhancing the locality. } 
  \label{fig:position} 
\end{figure}

The positional rearrangement procedure involves two key steps: clustering construction and matrix rearrangement, as illustrated in Figure \ref{fig:position}.
We begin by considering a set of $n$ assets that are initially arranged randomly within the corresponding {ADM}.
Financial data generally contains rich contextual information, including sector details, historical prices, volumes, etc., which can be used to divide the entire asset group into multiple distinct clusters.
Specifically, we propose a universal approach that uses hierarchical clustering based on dynamic time warping distance (\textit{DTW}) \cite{petitjean2011global} between assets' close prices for the cluster construction step. 
{DTW} enables flexible alignment and identification of similar patterns despite temporal variations, which makes it appropriate for financial price series exhibiting variations in terms of timing and speed.
In addition, hierarchical clustering provides the ability to reveal nested relationships at different levels of granularity, reflecting the varying degrees of similarity and hierarchical structures observed in financial assets based on factors such as sectors, industries, or market segments \cite{altomonte2013business}. In the second step, we sort assets based on the clustering outcomes, deriving new {ADM}s that represent rearrangements compared to the original {ADM}s.
Due to the hierarchical tree-like structure produced by agglomeration clustering, converting it into a sorted asset list is straightforward.
The complete procedure of positional rearrangement is illustrated in Algorithm \ref{algo:clustering}. 
Note that to enhance efficiency, we utilize coarse-grained monthly average return data when calculating the {DTW} distance.

\begin{algorithm}[htb]
\caption{ADM Positional Rearrangement}\label{algo:clustering}
\begin{algorithmic}[1]
\renewcommand{\algorithmicrequire}{\textbf{Input:}}
\renewcommand{\algorithmicensure}{\textbf{Output:}}
\REQUIRE A random list of $n$ assets $[a_1, a_2, ..., a_{n-1}, a_{n}]$, each asset's monthly average return series $\overline{r}_i$ in training part.
\ENSURE  A sorted list of $n$ assets $[a_1', a_2', ..., a_{n-1}', a_{n}']$, and the corresponding $\mathcal{M}_t$ over all the time
\STATE Compute the distance matrix $D$ based on the pairwise {DTW} distance between $\overline{r}_i$. 
\STATE Initialize each asset $a_i$ as a separate cluster. Create an ordered asset list $[a_i]$ for each cluster.
\WHILE{number of clusters $> 1$ }
\STATE Compute the pairwise distances between all clusters based on $D$ using the average linkage criterion.
\STATE Find the two closest clusters. 
\STATE Merge the two closest clusters into a new cluster.
\STATE Update the cluster label. Create a new ordered asset list by concatenating the ordered asset list of the two merged clusters.
\ENDWHILE
\STATE Return the ordered asset list $l_c$ of the top-level cluster. 
\STATE Generate {ADM}s over all the time based on $l_c$.
\end{algorithmic}
\end{algorithm}

\subsection{MoE Transformation Block} 
\label{sec:transform_block}
Despite the effective enhancement of the {ADM}'s locality property through the positional rearrangement, it does not fully capture the time-varying patterns of underlying assets' dependencies. 
A prime illustration is sector rotation driven by changing market conditions, economic factors, and investor sentiment. Sector rotation in finance refers to the cyclical pattern in which money flows between different sectors of assets. This concept is based on the idea that not all sectors of the economy perform equally well at the same time, due to varying economic conditions. When a particular sector comes into favor, the increased focus and investment can lead to assets within that sector moving more in unison leading to higher correlation. When the capital flows out of the sector, the assets return to normal correlation level. Thus, the dynamic and flexible end-to-end transformation is required to effectively capture the evolving dependencies.

\subsubsection{Quadratic Approximation}
\label{subsec:mlp-convlstm}
Given the universal approximation theorem \cite{SCARSELLI199815}, it is intuitive to consider the fully connected network (\textit{FCN}) as the transformation function. However, due to the large size of {ADM}s and the high-order nature of the transformation, achieving convergence with {FCN} can be challenging.
Therefore, we propose using the polynomial function to approximate the optimal transformation function \cite{eden1986polynomial, jacquin1993fractal}. Specifically, we adopt the quadratic function, which is widely recognized as an effective solution in function approximation problems \cite{922461, 4392762}. 
The quadratic function strikes a balance between capturing linear and high-order dependencies, making it well-suited for modeling the intricate complexity of financial markets with less risk of over-fitting. 
The quadratic function performs an element-wise transformation by assigning a new value to each entry of the {ADM}, infusing its ability to approximate both position rearrangement and data scaling. 
The formula of the quadratic transformation function can be written as follows:
\begin{equation}
    \label{eqn:trans}
    \mathbb{T}(\mathcal{M}_t^{k,u}) = \alpha_t \circ \mathcal{M}_t^{k,u} \circ \mathcal{M}_t^{k,u} + \beta_t \circ \mathcal{M}_t^{k,u} + \gamma_t 
\end{equation}
where $\alpha_t$, $\beta_t$, and $\gamma_t$ denote, respectively, the coefficient tensor of the quadratic term, the linear term, and the constant term, all of which have the shape $R^{1 \times c \times w \times h}$. ‘$\circ$’ denotes the Hadamard product. $\alpha_t = f_{\alpha}(\mathcal{M}_t^{k,u}|\Theta_\alpha)$, where $f_{\alpha}$ denotes the coefficient function outputting $\alpha_t$ given parameter set $\Theta_{\alpha}$ and input $\mathcal{M}_t^{k,u}$. Coefficients $\beta_t, \gamma_t$ are output by functions $f_{\beta}$ and $f_{\gamma}$ similarly.

We learn the coefficient functions $f_a$, $f_b$, and $f_c$ from the input {ADM} by designing proper neural architectures. 
To enhance the feature extraction from the input {ADM} sequence and reduce dimensionality, we apply two consecutive 3D convolutional layers \cite{Goodfellow-et-al-2016} to extract the latent feature. 
\begin{equation}
\label{equ:2conv}
     \boldsymbol{\phi}_t^{k,u} = W_{conv2} \ast (W_{conv1} \ast \mathcal{M}_t^{k,u} )
\end{equation}
where [$W_{conv1}$, $W_{conv2}$] denote the 3D convolution layers, and ‘$\ast$’ denotes the convolution operator.
The resulting output is subsequently passed to the following NN blocks, where each block learns its respective coefficient function. 
A straightforward way to model the coefficient function $f$ is by employing the Multilayer Perceptron (\textit{MLP}) \cite{Goodfellow-et-al-2016}
However, asset correlation could be influenced by various market factors such as macro-economy, interest rates, industry rotation, political events, etc. 
While {MLP} offers a simple means of approximating the coefficients, it fails to encode the complex market factors, leading to generalized rules that may not be suitable for all market scenarios. 
For example, in the presence of $K$ different market states caused by different market factors/events, the dependency between two stocks may vary across these states. 
Using a single {MLP} may result in a model that applies uniformly to all $K$ market situations, often leading to over-generalization of the model and suboptimal performance.

\begin{figure}[tb]
  \centering
  \setlength{\abovedisplayskip}{0pt}
  \setlength{\belowdisplayskip}{-10pt}
  \includegraphics[width=0.6\textwidth]{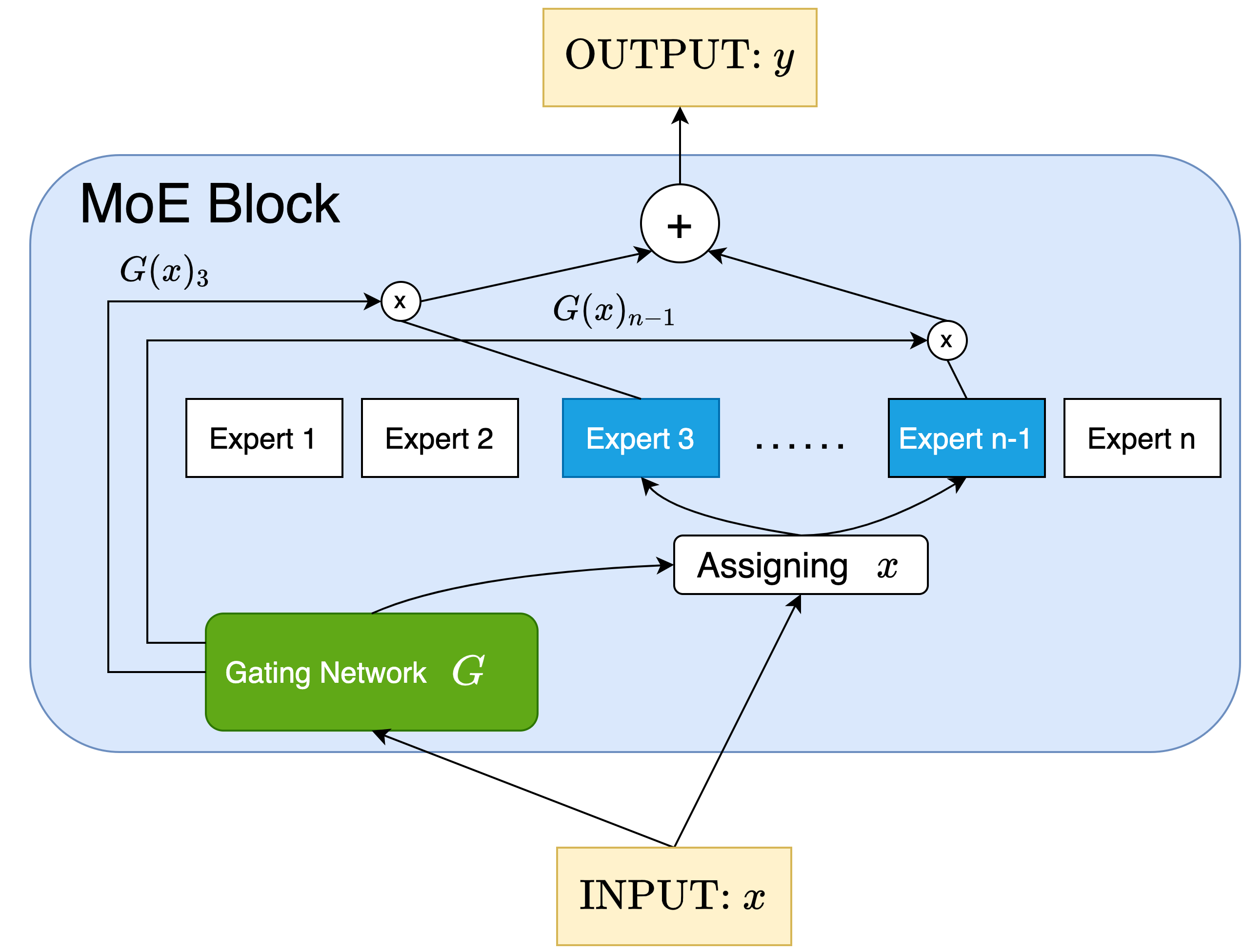} 
  \caption{{MoE} block structure. Assuming there are $n$ experts in the {MoE} block, the input is fed to the top 2 experts chosen by the gating network. } 
  \label{fig:MoE} 
\end{figure}

\subsubsection{Mixture of Expert Transformation}
\label{subsec:moe-convlstm}
Inspired by Mixture of Experts (\textit{MoE}) \cite{jacobs1991adaptive}'s success in solving complex data problems with multiple data regimes in the dataset \cite{eigen2013learning,shazeer2017outrageously,ma2018modeling}, we apply {MoE} to the transformation block for learning the transformation function $\mathbb{T}$ that can distinguish different market factors and transform input {ADM} to proper representation. A {MoE} block comprises a gating network and multiple subnets, each of which is an {MLP} (expert). It can achieve improvement, especially for complex datasets due to the gating network's capability of selecting a sparse combination of experts to process inputs with different data patterns. In our case, the gating network in {MoE} acts as a knowledgeable manager to decide which experts should be assigned the task of transforming $\mathcal{M}_t^{k,u}$ and combine their results afterward. The critical difference between {MLP} and {MoE} in constructing $\mathbb{T}$ is that {MLP} learns a single rule that generalizes all the complex market factors while {MoE} introduces sparse combinations of experts that specialize in individual market factor(s) and unify their outputs to generate a better {ADM} representation. 

The structure of an {MoE} block can be summarized as follows. $G(x)$ denotes The output of the gating network, and $E_i(x)$ denotes the output of the $i$-th expert network for a given input $x$, where $x =  \boldsymbol{\phi}_t^{k,u}$ (see Equation \ref{equ:2conv}) in our setting. 
To ensure load balancing, a tunable Gaussian noise, controlled by a trainable weight matrix $W_{noise}$, is introduced to encourage the gated network to pick different sets of experts.
Sparsity is achieved by selecting only the top $K$ expert values using the gating function \cite{716791} with a trainable weight matrix $W_{g}$. 
The output $y$ of the {MoE} module is calculated by:
\begin{equation}
\begin{aligned}
    y &= \sum\limits_{i=1}^n G(x)_{i} E_{i}(x)\\
    G(x) &= \mbox{Softmax}(\mbox{TopK}(H(x), k))\\
    H(x)_{i} &= (x \cdot W_{g})_{i} + \mbox{SNorm}() \cdot \mbox{Softplus}((x \cdot W_{noise})_{i})\\
    \mbox{TopK}(v, k)_{i} &= \left\{
    \begin{array}{l}
    v_{i} \hspace{0.2cm}  \mbox{if $v_{i}$} \in \mbox{top k elements of v}
    \\
    - \infty \hspace{0.2cm}  \mbox{ otherwise}
    \end{array}
    \right\}
\end{aligned}
\label{equ:MoE}
\end{equation}

The sparsely-gated feature is an indispensable component to the improvement of prediction accuracy, as selecting only a sparse set of experts each time ensures less noise and over-generalization of each expert.
Figure \ref{fig:MoE} illustrates the idea. 
Given $n$ experts, this design increases the learning capacity of the transformation function by offering $n \choose 2$ different combinations of experts, whereas an MLP can only learn one generalized rule for all inputs. Moreover, it aids in noise reduction as each expert can focus on its specialized area. 

In summary, we employ three separate {MoE} blocks to model the coefficient functions $f_{\alpha}$, $f_{\beta}$, and $f_{\gamma}$ used in the quadratic approximation (see Equation \ref{eqn:trans}), shown as follows:  
\begin{equation}
\label{equ:transform}
\begin{aligned}
    \alpha_t &=  \mbox{MoE}_{quad}(\boldsymbol{\phi}_t^{k,u}|\Theta_{\alpha})\\
    \beta_t &=  \mbox{MoE}_{linear}(\boldsymbol{\phi}_t^{k,u}|\Theta_{\beta})\\ 
    \gamma_t &=  \mbox{MoE}_{const}(\boldsymbol{\phi}_t^{k,u}|\Theta_{\gamma})\\ 
    \tilde{\mathcal{M}}_t^{k,u} &= \alpha_t \circ \mathcal{M}_t^{k,u}  \circ \mathcal{M}_t^{k,u}  + \beta_t \circ \mathcal{M}_t^{k,u} + \gamma_t
\end{aligned}
\end{equation}

\subsection{Spatiotemporal Block} 
\label{subsec:convlstm}
After passing through the transformation block introduced in Section \ref{sec:transform_block}, the transformed {ADM} sequence is sent to the spatiotemporal block for the extraction of the spatial and temporal signals encoded in $\tilde{\mathcal{M}}_t^{k,u}$. 
Specifically, we adopt the \textit{ConvLSTM} \cite{shi2015convolutional} as the spatiotemporal block due to its superior performance in spatiotemporal-related tasks. The core unit of the {ConvLSTM} network is the {ConvLSTM} cell. The input and intermediate states fed to the cell are tensors with the shape $R^{c \times w \times h}$. Denote the input state as $X_t$, memory state as $C_t$, hidden state as $H_t$, and gated signals as $i_t$, $f_t$, $g_t$, $o_t$ at each timestamp of one cell. The future state of a certain {ADM} entry in the matrix is determined by the inputs and past states of its local neighbors. This can easily be achieved by using a convolution operator in state-to-state and input-to-state transitions:
\begin{equation}
\begin{aligned}
    i_t &= \sigma(W_{xi} \ast  X_t + W_{hi} \ast \mathcal{H}_{t - 1} + W_{ci} \circ C_{t - 1} + b_i)
    \\f_t &= \sigma(W_{xf} \ast X_t + W_{hf} \ast \mathcal{H}_{t - 1} + W_{cf} \circ C_{t - 1} + b_f )
    \\C_t &=  f_t \circ C_{t - 1} + i_t \circ tanh(W_{xc} \ast X_t + W_{hc} \ast \mathcal{H}_{t - 1} + b_c)
    \\o_t &= \sigma(W_{xo} \ast X_t + W_{ho} \ast \mathcal{H}_{t - 1} + W_{co} \circ C_t + b_o)
    \\\mathcal{H}_t &= o_t \circ tanh(C_t)
\end{aligned}
\end{equation}

\subsection{Loss Function}
We employ the mean square error (\textit{MSE}) loss to optimize model parameters. 
Denote the output matrix of the spatiotemporal block as $\mathcal{E}_{t+h}$.
Considering the inherent symmetry of {ADM}, we confine the loss calculation to the upper triangular part of $\mathcal{E}_{t+h}$ to reduce computational complexity and redundancy. The MSE loss is then computed as follows: 
\begin{equation}
\begin{aligned}
    L_{sqr} = (\mathcal{M}_{t+h}-\mathcal{E}_{t+h})^2 = \sum_{i=1}^{n} \sum_{j=i}^{n} (f(a_i,a_j)_{t+h} - \hat{f}(a_i,a_j)_{t+h})^2
\end{aligned}
\end{equation}
where $\hat{f}$ denotes the predicted dependency measurement of two assets $a_i$ and $a_j$.
To generate the forecasted {ADM}, we construct a symmetric matrix $\hat{\mathcal{M}}_{t+h}$ by duplicating the values from the upper triangular part of $\mathcal{E}_{t+h}$ as follows:
\begin{equation}
    \hat{\mathcal{M}}_{t+h}[i, j] = \left\{
    \begin{array}{l}
    \mathcal{E}_{t+h}[i, j], \hspace{0.2cm}  i \leq j\\
    \mathcal{E}_{t+h}[j, i], \hspace{0.2cm}  i > j
    \end{array}
    \right\}
\end{equation}

\subsection{PSD Enforcement Block}
Positive semidefiniteness (\textit{PSD}) is a crucial property of {ADM} where all of its eigenvalues are non-negative, which ensures mathematical consistency, meaningful interpretations, and reliable subsequent applications. For example, consider portfolio variance denoted by Equation \ref{equ:psd} where $w_i$ represents the weight of asset $i$ in the portfolio. If the covariance matrix fails to meet the {PSD} condition, the resulting portfolio will possibly incur a negative variance. This implies a constraint that the covariance matrix $Cov$ must be positive semidefinite. 

\begin{equation}
\begin{aligned}
    \operatorname{var}\left(\sum_i w_i X_i\right)=&\sum_i \sum_j w_i w_j Cov_{i, j}
\end{aligned}
\label{equ:psd}
\end{equation}

The {ADM} forecasted by the above structure does not inherently guarantee positive semi-definiteness.
To address this limitation, we introduce a {PSD} enforcement block. 
We obtain the nearly identical {PSD} matrix $\hat{\mathcal{M}}_{t+h}^*$ by setting negative eigenvalues of $\hat{\mathcal{M}}_{t+h}$ to zero \cite{XU2021110072}, shown as follows:

\begin{equation}
\begin{aligned}
&\hat{\mathcal{M}}_{t+h} =  \mathcal{P} \Lambda \mathcal{P}^T\\
&\hat{\Lambda} = \text{{diag}}(\max(\Lambda, 0)) \\
&\hat{\mathcal{M}}_{t+h}^* = \mathcal{P} \hat{\Lambda} \mathcal{P}^T
\end{aligned}
\end{equation}
where $\mathcal{P}$ is an orthogonal matrix of eigenvectors and $\Lambda$ is a diagonal matrix of eigenvalues. 
Even though this approach does not guarantee finding the exact closest {PSD} matrix, it efficiently leverages the symmetry and near {PSD} property of matrix $\hat{\mathcal{M}}_{t+h}$ to approximate the closest {PSD} matrix.

\section{Experiment}\label{sec:experiment}
In the experiment, we assess the performance of the proposed framework using real-world datasets and conduct a comprehensive analysis of the model behaviors.

\subsection{Experimental Settings}

\subsubsection{Datasets}\label{subsec:datasets}
We construct a pool of real-world stock data from S\&P-100, NASDAQ-100, and DJI-30, encompassing the most impactful companies over the past 15 years (from 2005/09/27 to 2020/08/05). 
We retain stocks with complete closing price data throughout the whole period, resulting in 133 stocks in the pool, each of which has a time series of 3740 price data points. 
The {ADM} sequences can be generated from the stock price data as depicted in Figure \ref{fig:adm_structure}.
A general guideline from finance is to choose the time lag $n_{lag}$ at least as large as the number of assets $n$ to avoid the out-of-sample problem \cite{tashman2000out}. However, if $n_{lag}$ is too large, the model may be incapable of capturing short-term (e.g. monthly) dynamics of the {ADM} sequences.
To strike a balance, we set $n_{lag}=42$ and $n=32$. 
Without the loss of generality, we construct $10$ stock datasets by randomly picking stocks from the pool, each of which contains $32$ stocks (see Appendix for detailed asset components).
In each dataset, we first obtain $3740 - 1$ = $3739$ log return data points. By setting $n_{lag} = 42$, we generate $3739 - n_{lag} + 1$ = $3698$ {ADM}s.
In this paper, we mainly study the {ADM}s' monthly evolution. We set $h = u = 21$, $k = 10$, resulting in the moving window with length $u * k+1$ = $211$.
As the rolling windows move forward each day, we generate $3698 - 211 + 1$ = $3488$ data samples.
We select the first $90\%$ of {ADM} sequences as training samples (including validation) and the remaining $10\%$ as testing samples.
The model performance is evaluated on each stock dataset independently to guarantee the generosity of the experimental outcome.

\subsubsection{Evaluation Metrics}\label{subsec:metrics}
We evaluate {ADM} prediction accuracy using two metrics: \textbf{mean square error (MSE)} and \textbf{gain}. 
The \textit{MSE} metric measures the average squared difference between the predicted {ADM}s and the groundtruth {ADM}s. 
Given the symmetry of {ADM}s, we use only the upper triangular part for computation.
The \textit{Gain} metric quantifies the improvement in prediction accuracy compared to the conventional practice of using the previous {ADM} as the future {ADM} in finance \cite{elton1978betas}. A larger gain indicates better performance over the conventional approach. 
Denote the i-th forecasted {ADM} as $\hat{\mathcal{M}_i}$, groundtruth {ADM} as $\mathcal{M}_i$, and the previous {ADM} as $\mathcal{M}_{i-h}$. Formally, the evaluation metrics are computed by:
\begin{equation*}
\begin{aligned}
    \mbox{MSE} = \frac{1}{N} \sum_{i=1}^N (\mathcal{M}_{i} - \hat{\mathcal{M}}_{i})^2,~
    \mbox{Gain} = 1 - \frac{\sum_{i=1}^N (\mathcal{M}_{i} - \hat{\mathcal{M}}_{i})^2}{\sum_{i=1}^N (\mathcal{M}_{i} - \mathcal{M}_{i-h})^2}
\end{aligned}
\end{equation*}

\subsubsection{Baselines}\label{subsec:baselines}
We compare the proposed framework to six baselines covering two classical statistical models for correlation forecasting and four spatiotemporal deep learning models specifically tailored for the sequential forecasting of {ADM}
For fairness, we intentionally exclude methods that incorporate external information such as interest rates and financial indexes from the comparison.
The baseline models are listed as follows.
\begin{itemize}
    \item{\textbf{Constant Correlation Model (CCM)}} is a statistical model that considers the correlation of returns between any pair of securities to be the same \cite{elton1973estimating,elton1978betas}. It takes the historical means of correlation as the prediction. 
    \item{\textbf{DCC-Garch}\cite{engle2002dynamic}} is a statistical model that incorporates the dynamic nature of correlation by parameterizing the conditional correlations. It forecasts the future correlation matrix and utilizes the predicted correlation matrix to generate covariance matrices.
    \item{\textbf{Long-Short-Term-Memory network (LSTM)}} \cite{HochSchm97} is a widely used deep learning model for sequence prediction. In our approach, we flatten each input {ADM} into a 1D vector and use a standard {LSTM} implementation.  
    \item{\textbf{Raw-ConvLSTM}} \cite{shi2015convolutional} directly feeds the raw {ADM} sequences to the convolutional {LSTM} network (\textit{ConvLSTM}) without any preprocessing or transformation.
    \item{\textbf{FCN-ConvLSTM}} is an end-to-end framework based on {ConvLSTM}. It incorporates a fully connected network (\textit{FCN}) to transform the input {ADM}, which is then passed to the {ConvLSTM} for the prediction task.  
    \item{\textbf{Convolution-3D (Conv3D)}} \cite{8594916} is a deep learning model with an encoder-decoder architecture. The encoder and decoder contain three consecutive 3D-convolution layers respectively.  
\end{itemize}

\begin{table}[t]
\centering
\caption{Parameters Pool.}
\footnotesize
\begin{tabular}{lll}
\hline
Name&Range&Optimal\\
\hline
\textbf{Common}&\\
k&\{2,3,6,10,12\}&10\\
$\mathrm{size}_c$&\{3,5\}&5\\
optimizer&\{Adam, SGD\}&Adam\\
grad\_clip&\{1,5,10\}&10\\
\hline
\textbf{ADNN}&\\
$\mathrm{size}_b$&\{64, 128, 256, 384, 512\}&128\\
init\_lr&\{1e-2, 5e-3, 1e-3, 5e-4\}&5e-4\\
$n_{exp}$&\{2, 4, 8, 16\}&8\\
\multirow{4}*{$\mathrm{top}_k$}&\{1,2\} if $n_{exp} = 2$\\
&\{1,2,3,4\} if $n_{exp} = 4$ \\
&\{1,2,4,8\} if $n_{exp} = 8$ & 4\\
&\{1,4,8,16\} if $n_{exp} = 16$\\
\hline
\textbf{FCN-ConvLSTM}&\\
$\mathrm{size}_b$&\{128, 256, 384, 512, 640\} & 256\\
init\_lr&\{1e-2, 5e-3, 1e-3, 5e-4\} & 1e-3\\
$n_l$&\{2,3,4\}&3\\
\hline
\end{tabular}
\label{tab:hyper}
\end{table}

\subsubsection{Environment and Parameter Tuning}
We run the experiment with one NVIDIA GeForce RTX 3090 Graphic Card and use the Adam optimizer \cite{kingma2014adam} to train deep learning models.  
We utilize the gradual warm-up learning rate scheduler \cite{goyal2017accurate} and the next scheduler reduces the learning rate when the validation loss stops decreasing over a certain number of epochs.
For each baseline model and the proposed framework, we repeatedly run the model with various groups of hyperparameters and use the Bayesian method to fine-tune the proposed framework. 

The main parameters and their corresponding range are shown in Table \ref{tab:hyper}. 
\textit{Common} refers to parameters across all the methods, where $k$ denotes the length of the input sequence, $\mathrm{size}_c$ denotes the filter size of the convolution kernel in the ConvLSTM, \textit{grad\_clip} denotes the value for gradient clipping, 
$\mathrm{size}_b$ denotes the batch size, and \textit{init\_lr} denotes the initial learning rate for the adaptive learning rate scheduler. 
For parameters that are specific to \textit{ADNN}, \textit{degree} denotes the degree of the transformation function as shown in Equation \ref{equ:transform}, $n_{exp}$ denotes the number of experts in the \textit{MoE} network, and $\mathrm{top}_k$ denotes the number of experts participating in the generation of the result. 
For \textit{FCN-ConvLSTM}, $n_l$ denotes the number of layers of \textit{FCN}. 
The \textit{Optimal} column shows the tuned optimal parameters for the experiment. For the tuned parameters listed under \textit{Common}, all the methods achieve the best result at the reported value. Notice that $\mathrm{size}_c$ will not affect \textit{Conv3d} and \textit{LSTM}. 
Some parameters are not included in Table \ref{tab:hyper}. For \textit{DCC} and \textit{CCM}, we use a rolling window of size $5$ years. For \textit{Conv3d}, the encoder/decoder contains three consecutive 3D-convolution layers. For \textit{FCN-ConvLSTM}, the hidden size of \textit{FCN} is twice the input size. Horizon $h$ is an application-specific parameter, and since our application is portfolio management with monthly portfolio adjustments, we set $h=21$.

\subsection{Evaluation of {ADM} Prediction Accuracy}\label{subsec:eval_accuracy}

\begin{table}[tb]
\centering
\caption{Performance Comparison on Different Datasets}
\footnotesize
\setlength{\tabcolsep}{3.5pt}
\begin{tabular}{l|l|lllll}
\hline
\multicolumn{2}{c|}{\diagbox{Method}{Dataset}}&Set 1&Set 2&Set 3&Set 4&Set 5\\ \hline\hline
\multirow{2}*{CCM}&MSE&0.0613&0.0609&0.0673&0.0686&0.0754\\\cline{2-7}
{}&Gain&-0.342&-0.314&-0.431&-0.532&-0.534\\\hline
\multirow{2}*{DCC}&MSE&0.0643&0.0647&0.0609&0.0600&0.0639\\\cline{2-7}
{}&Gain&-0.405&-0.398&-0.293&-0.339&-0.30\\\hline
\multirow{2}*{Conv3D}&MSE&0.0630&0.0643&0.0647&0.0549&0.0583\\\cline{2-7}
{}&Gain&-0.379&-0.392&-0.374&-0.222&-0.187\\\hline
\multirow{2}*{LSTM}&MSE&0.8346&0.9155&0.7920&0.8303&0.8261\\\cline{2-7}
{}&Gain&-17.3&-18.8&-15.8&-15.8&-15.8\\\hline
{Raw-Conv}&MSE&0.0360&0.0359&0.0378&0.0364&0.0403\\\cline{2-7}
{LSTM}&Gain&0.210&0.220&0.199&0.188&0.179\\\hline
{FCN-Conv}&MSE&0.0581&0.0411&0.0596&0.0571&0.4939\\\cline{2-7}
{LSTM}&Gain&-0.275&0.111&-0.266&-0.270&-0.009\\ \hline\hline
\multirow{2}*{P-ADNN}&MSE&0.0351&\textbf{0.0353}&0.0376&\textbf{0.0334}&0.0368\\\cline{2-7}
&Gain&0.224&\textbf{0.221}&0.190&\textbf{0.247}&0.246\\\hline
\multirow{2}*{T-ADNN}&MSE&0.0349&0.0358&\textbf{0.0367}&0.0360&0.0371\\\cline{2-7}
&Gain&0.228&0.210&0.209&0.198&0.221\\\hline
\multirow{2}*{PT-ADNN}&MSE&\textbf{0.0336}&0.0356&0.0368&0.0367&\textbf{0.0363}\\\cline{2-7}
&Gain&\textbf{0.256}&0.214&\textbf{0.207}&0.171&\textbf{0.256}\\\hline
\end{tabular}
\label{tab:group}
\end{table}

\begin{table}[!htb]
\ContinuedFloat
\centering
\caption{Performance Comparison on Different Datasets, continued}
\footnotesize
\setlength{\tabcolsep}{3.5pt}
\begin{tabular}{l|l|lllll|l}
\hline
\multicolumn{2}{c|}{\diagbox{Method}{Dataset}}&Set 6&Set 7&Set 8&Set 9&Set 10&Mean(Std)\\ \hline\hline

\multirow{2}*{CCM}&MSE&0.0686&0.06&0.0605&0.0724&0.0626&0.0657(0.0052)\\\cline{2-8}
{}&Gain&-0.455&-0.381&-0.385&-0.709&-0.457&-0.454(0.1097)\\\hline
\multirow{2}*{DCC}&MSE&0.0669&0.0566&0.0660&0.0630&0.0609&0.0627(0.0029)\\\cline{2-8}
{}&Gain&-0.416&-0.303&-0.514&-0.482&-0.424&-0.388(0.0730)\\\hline
\multirow{2}*{Conv3D}&MSE&0.0579&0.0566&0.0605&0.0605&0.0605&0.0601(0.0031)\\\cline{2-8}
{}&Gain&-0.231&-0.302&-0.387&-0.421&-0.414&-0.331(0.0833)\\\hline
\multirow{2}*{LSTM}&MSE&0.8048&0.6813&0.7707&0.8388&0.8261&0.8120(0.0565)\\ \cline{2-8}
{}&Gain&-16.0&-14.6&-16.7&-18.7&-18.3&-16.9(1.3637)\\\hline
{Raw-Conv}&MSE&0.0376&0.0647&0.0334&0.0336&0.0329&0.0389(0.0089)\\\cline{2-8}
{LSTM}&Gain&0.202&-0.487&0.236&0.211&0.234&0.140(0.2096)\\\hline
{FCN-Conv}&MSE&0.0626&0.0402&0.0579&0.0592&0.0562&0.0986(0.132)\\\cline{2-8}
{LSTM}&Gain&-0.329&0.077&-0.325&-0.398&-0.307&-0.199(0.1754)\\\hline\hline
\multirow{2}*{P-ADNN}&MSE&0.0379&0.0323&0.0322&0.0361&0.0342&0.0351(0.0019)\\\cline{2-8}
&Gain&0.183&0.258&0.254&0.130&0.193&0.215(0.0385)\\\hline
\multirow{2}*{T-ADNN}&MSE&0.0358&\textbf{0.0321}&0.0320&0.0336&0.0321&0.0346(0.0019) \\\cline{2-8}
&Gain&0.226&\textbf{0.262}&\textbf{0.263}&0.218&0.245&0.228(0.0217) \\\hline
\multirow{2}*{PT-ADNN}&MSE&\textbf{0.0337}&0.0328&\textbf{0.0319}&\textbf{0.0324}&\textbf{0.0321}&\textbf{0.0342(0.0019)} \\\cline{2-8}
&Gain&\textbf{0.273}&0.245&0.262&\textbf{0.220}&\textbf{0.245}&\textbf{0.235(0.0297)} \\\hline
\end{tabular}
\label{tab:group}
\end{table}

\begin{table}[t]
\centering
\footnotesize
\caption{Comparison of Different Positional Rearrangement Approach}
\begin{tabular}{c|c||c|c}
\hline
\multicolumn{2}{c||}{\diagbox{Framework}{Metrics}} & { Avg. MSE}& { Avg. Gain} \\ \hline
\multicolumn{2}{c||}{K-means}& 0.0371 & 0.181 \\ \hline
\multirow{2}*{Hierachy}&AutoCorr& 0.0385 & 0.160 \\ \cline{2-4}
&DTW& \textbf{0.0351} &  \textbf{0.215}\\ \hline

\end{tabular}
\label{tab:ablation_2}
\end{table}

The overall prediction accuracy result is shown in Table \ref{tab:group}. 
Here, \textit{P-ADNN} refers to the framework in Figure \ref{fig:general_framework} with only positional rearrangement block and without the MoE transformation block, \textit{T-ADNN} refers to the framework with only {MoE} transformation and without the positional rearrangement, and \textit{PT-ADNN} integrates both forms of transformation in sequential order.
We can observe deep learning-based methods generally outperform statistical methods (\textit{CCM} and \textit{DCC}), indicating the superiority of data-oriented methods over methods that greatly depend on unrealistic statistical assumptions in the complex and dynamic financial market. 
\textit{LSTM}/\textit{Conv3D}, which are simple models that utilize only temporal/spatial information, yields unsatisfactory prediction accuracy. 
The inferior performance of \textit{LSTM} further demonstrates the importance of including spatial signal in {ADM} prediction task compared to considering temporal signal only. 
By incorporating both spatial and temporal information, \textit{ConvLSTM} improves prediction accuracy. To study if the addition of the transformation layer can help alleviate the limit imposed by the {ADM} representation problem on raw \textit{ConvLSTM}, we evaluate the \textit{FCN-ConvLSTM} and the more complex \textit{ADNN} models. 
We can see that the performance improvement of \textit{FCN-ConvLSTM} is marginal and sometimes even negative. Even though \textit{FCN-ConvLSTM} uses an end-to-end architecture that should learn better representation than raw inputs in principle, it displays worse performance, which indicates that without a proper design of the transformation, the utilization of deep layers cannot boost the performance.

After comparing multiple \textit{ADNN} frameworks with different transformation functions, we can observe that \textit{PT-ADNN} that utilizes both positional rearrangement and {MoE} transformation achieves the best performance in most cases.
All the proposed \textit{ADNN} frameworks exhibit superior performance compared to the \textit{Raw-ConvLSTM} without using any transformation, validating the necessity of a well-designed transformation function for the raw {ADM} to alleviate the {ADM} representation problem.
In specific cases (e.g. set 2 and set 4), the \textit{P-ADNN}, which solely utilizes positional rearrangement, exhibits superior performance. 
This observation can be ascribed to the stable patterns and limited variability present in the dataset. In such cases, complex transformations might be less effective as they could introduce unnecessary intricacies or capture noise, potentially resulting in overfitting.
In other cases (e.g. set 7 and set 8), the \textit{T-ADNN} using solely the {MoE} transformation outperforms others, which may be due to the absence of clear grouping behavior among assets within the group, which diminishes the effectiveness of similarity-based positional rearrangement.
In conclusion, from Table \ref{tab:group} we can see \textit{ADNN} has consistent performance advantages on different datasets over various baselines.  

\subsection{Ablation Study}\label{subsec:ablation}
\subsubsection{Comparison of Positional Rearrangement Methods}

In Section \ref{subsec:position}, we mention using the hierarchical clustering approach based on the \textit{DTW} distance. To further validate the effectiveness of the proposed method, we compare this design to two other kinds of clustering methods: (1) The \textit{K-means} approach constructs clusters by using the monthly returns series as the feature vector. (2) The hierarchical clustering approach uses autocorrelation as the distance metric. Table \ref{tab:ablation_2} displays the results.
We can observe that hierarchical clustering using \textit{DTW} achieves the best performance, which validates its efficacy in improving the locality property of {ADM}, thereby augmenting the performance of the spatiotemporal block. 
The performance improvement is likely attributed to the inherent structure of the financial market, where assets are organized into industrial sectors. Mutual and hedge funds typically engage in investment activities by tracking industrial indices. This practice results in assets within the same sector being influenced simultaneously, leading to a high correlation among them, making hierarchical clustering a good choice.

\subsubsection{Comparison of Different Expert Numbers}\label{subsec:influence_n_exp}

\begin{table*}[tbp]
\centering
\caption{Influence of $n_{exp}$ and $top_k$ on Prediction {MSE}.}
\footnotesize
\begin{tabular}{c|cccc}
\hline
\diagbox[height=1.6\line]{\textbf{Set}}{$\mathbf{n_{exp}}$} & 2 ($\textrm{top}_k = 1$) & 4 ($\textrm{top}_k = 2$) & 8 ($\textrm{top}_k = 4$) & 16 ($\textrm{top}_k = 4$) \\
\hline
1 & 0.0575 & 0.0349 & 0.0349 & 0.0345 \\
2 & 0.0566 & 0.0370 & 0.0358 & 0.0383 \\
3 & 0.0579 & 0.0375 & 0.0367 & 0.0490 \\
4 & 0.0362 & 0.0362 & 0.0360 & 0.0370 \\
5 & 0.0613 & 0.0387 & 0.0371 & 0.0379 \\
6 & 0.0358 & 0.0498 & 0.0358 & 0.0366 \\
7 & 0.0532 & 0.0319 & 0.0321 & 0.0324 \\
8 & 0.0554 & 0.0332 & 0.0320 & 0.0336 \\
9 & 0.0554 & 0.0341 & 0.0336 & 0.0341 \\
10 & 0.0554 & 0.0353 & 0.0321 & 0.0315 \\
Avg. & 0.0525 & 0.0369 & \textbf{0.0346} & 0.0365 \\
\hline
\end{tabular}
\label{tab:moe}
\end{table*}

In Section \ref{sec:transform_block}, we introduce the {MoE} transformation. The performance of {MoE} depends on two crucial parameters, (1) the number of experts $n_{exp}$, which determines how many experts in total are contained in the network, and (2) $\mathrm{top}_k$, which determines how many experts participate in generating the final transformation function ($\mathrm{top}_k \leq n_{exp}$). Table \ref{tab:moe} shows how these two parameters affect the learning of the transformation function $\mathbb{T}$ and in turn the prediction {MSE}. The entry in the table denotes the {MSE} returned by the optimal $top_k$ on average in the 10 stock datasets for a given $n_{exp}$. For example, when $n_{exp}=8$, on average $top_k=4$ produces the optimal {MSE} among the 10 stock datasets. We have several observations from the table: (1) the optimal ($n_{exp}, top_k$) pair is (8,4), (2) The effectiveness of $n_{exp}$ and $top_k$ does not follow a simple "larger is better" pattern. This phenomenon may arise from the initial increase of $n_{exp}$ to an optimal number effectively capturing the complexities of market data. However, excessive increments in $n_{exp}$ impose risks on overfitting and detrimentally impact performance.

\subsubsection{Comparison of Polynomial Approximation Functions}
\begin{table}[t]
\centering
\footnotesize
\caption{Comparison of Different Polynomial Approximation Functions}
\begin{threeparttable}
\begin{tabular}{c|c||c|c}
\hline
\multicolumn{2}{c||}{\diagbox{Framework}{Metrics}} & { Avg. MSE}& { Avg. Gain} \\ \hline
\multirow{3}*{Matmul}&Linear& 0.0447 & 0.026 \\ \cline{2-4}
&Quadratic& 0.0413 & 0.100 \\\cline{2-4}
&Cubic& 0.0417 & 0.091\\ \hline
\multirow{3}*{Hadamard}&Linear& 0.0524 & 0.026 \\ \cline{2-4}
&Quadratic \tnote{a}& \textbf{0.0346} & \textbf{0.228} \\\cline{2-4}
&Cubic& 0.0417 & 0.091\\ \hline

\end{tabular}
\begin{tablenotes}
    \item[a] Used by {ADNN}.
  \end{tablenotes}
\end{threeparttable}
\label{tab:ablation_1}
\end{table}

To validate the efficacy of the polynomial approximation function proposed in Equation \ref{eqn:trans}, we conduct experiments involving alternative transformation functions of various degrees (linear and cubic) within the polynomial approximation function, as well as different matrix multiplication methods. 
We compare these approaches using the average {MSE} and {gain} performance across the ten datasets.
Table \ref{tab:ablation_1} summarizes the results. 
We can observe that the approximation function in quadratic form reaches the best performance in both matrix multiplication cases. 
At the same time, the quadratic approximation function with the Hadamard product reaches the best performance in terms of {MSE} and {gain}. 
These findings indicate the importance of a well-designed transformation function for accurate forecasting of future {ADM}.
Notably, an increase in degrees generally incurs higher costs and longer convergence times. 

\subsection{Explore MoE}
The financial market is characterized by its dynamic nature, as different market factors come into prominence during different time periods.
For instance, during a financial crisis, the dominating market factor is the fear of company bankruptcy due to the severe lack of liquidity. 
Consequently, most stocks experience a downturn with heightened correlations driven by the pessimistic macroeconomic sentiment.
Conversely, during normal market conditions, stock prices are primarily influenced by companies' fundamentals, resulting in relatively lower correlations among stocks from different industries. 
Moreover, market factors may intertwine, as macroeconomic conditions and company fundamentals might simultaneously exert their influence, resulting in correlations behaving in a sophisticated manner. 

In Section \ref{subsec:eval_accuracy}, we show that the {MoE} block can improve the model performance because of its capability to distinguish different market scenarios and thus better model the asset correlation dynamics. In this section, we investigate \textit{how} {MoE} distinguishes market scenarios and try to visualize the process. The intuition is as follows. If {MoE} can successfully distinguish different market factors, we would expect it to select similar expert combinations in the market phases dominated by the same market factors and different combinations of experts for different market phases dominated by different market factors. Since it is impossible to enumerate and quantify the market factors comprehensively in real-world financial markets, we choose to \textit{simulate} different market scenarios with synthetic market data, which is a common approach in the finance field. We divide a market scenario into several market phases $\{\mbox{MP}^j | j=1,2,...\}$, each of which is generated with some market factor(s). We denote market phase $\mbox{MP}^j$ that is solely generated from market factor $\mbox{MP}_i$ as $\mbox{MP}_i^j$. 

The stock price data in each market phase $\mbox{MP}^j$ are generated with the multidimensional geometric Brownian motion function (\textit{MGBM}), which is often used in Monte Carlo simulation in financial experiments \cite{glasserman2004monte}. {MGBM} requires three inputs, namely, correlation matrix $cm$, expected return $exp$ and volatility $vol$. The triplet $(cm_i, exp_i ,vol_i)$ forms a market factor $\mbox{MP}_i$, which is defined as follows:
\begin{itemize}
    \item Correlation Matrix ($cm_i$): a random correlation matrix is generated given a vector of eigenvalues following a numerically stable algorithm spelled out by \cite{davies2000numerically}. The random eigenvalues are generated from a uniform distribution with a lower boundary equal to 0.1 and an upper boundary equal to 2.
    \item Expected Return ($exp_i$): the value is randomly sampled from a uniform distribution with a lower boundary equal to 0.1 and an upper boundary equal to 0.5.
    \item Volatility ($vol_i$): the value is randomly sampled from a uniform distribution with a lower boundary equal to 0.01 and an upper boundary equal to 0.1.
\end{itemize}
The simulation of the stock price data for market phase $\mbox{MP}_i^j$ can be formulated with the following equation: 
\begin{equation}
    price(\mbox{MP}_i^j) = \mbox{MGBM}(cm_i, exp_i ,vol_i)
    \label{equ:price_mgbm}
\end{equation}
We create three different market scenarios (1 to 3) with different setups of the triplet, as outlined in Table~\ref{table:scparam}.
To save space, we focus our discussion solely on the quadratic-term {MoE} block (see $a_t$ in Equation \ref{equ:transform}).
We finetune the model parameters based on each corresponding scenario and visualize the behavior of {MoE} in the optimal setting.

\begin{figure*}[thb]
    \centering
    \subfloat[Scenario 1.]{\includegraphics[width=0.27\textwidth]{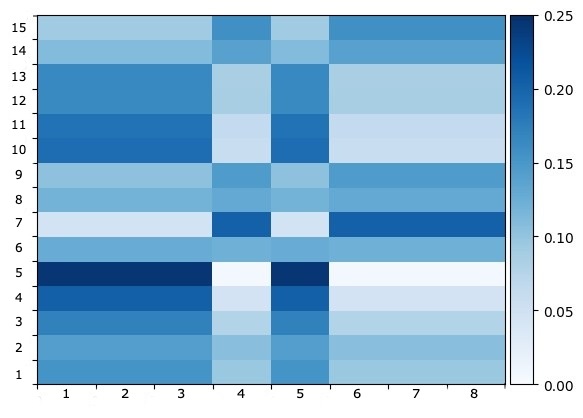}}
    \hspace{2em}%
    \subfloat[Scenario 2.]{\includegraphics[width=0.27\textwidth]{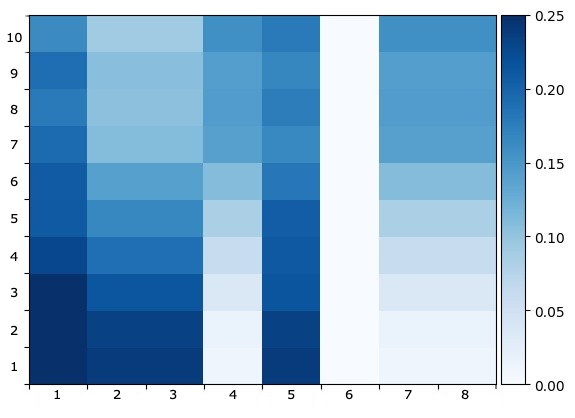}}
    \hspace{2em}%
    \subfloat[Scenario 3.]{\includegraphics[width=0.3\textwidth]{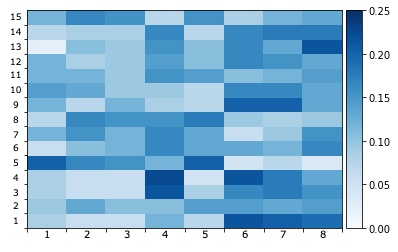}}
    \caption{{MoE} visualization under different market scenarios.}
    \label{fig:moe_visual}
\end{figure*}

Figure \ref{fig:moe_visual} depicts the chance that the experts are selected across different market scenarios, where the $y$-axis denotes the market phases and the $x$-axis represents the experts in {MoE}.
The discrete $x$ and $y$ axes divide the figure into grids and the color of a grid denotes the chance of the expert being chosen in a market phase. 
Given a market phase $\mbox{MP}^j$ comprising $N$ input {ADM}s sequences, the importance weight of $exp_i$ for $\mbox{MP}_j$ is calculated by: 
\begin{equation}
\begin{aligned}
    w(exp_i^j) &= \frac{1}{N}\sum\limits_{n=1}^N \mathbf{1}_{j,k}^n (exp_i)
    \label{eqn:expert_weight}\\
    \mathbf{1}_{j,k} (exp_i) &= \left\{
    \begin{array}{l}
    \frac{1}{top_k} \hspace{0.2cm}  \mbox{if $exp_{i}$ is selected for $\mbox{MP}_j$}
    \\
    0 \hspace{0.2cm}  \mbox{ otherwise}
    \end{array}
    \right\}
\end{aligned}
\end{equation}
For instance, in Scenario 1 where $n_{exp} = 8$ and $top_k = 4$, the weight assigned to a selected expert is $\frac{1}{top_k} = 0.25$. 
If this expert is chosen for all $N$ input sequences within this market phase, it has average weight $w(exp_i^j)=0.25$. 
Notably, the maximum value of any $w(exp_i^j)$ is $\frac{1}{top_k}$, while the smallest value would be zero if the expert is not selected at all.

\begin{table}
\centering
\caption{{MoE} Parameters under Different Market Scenarios.}
\footnotesize
\begin{tabular}{l|l|l|l}
\hline
 Scenario & 1 & 2 & 3 \\ \hline\hline
 No. Market Phases & 15 & 10 & 15 \\ \hline
 Trading Days per Phase & 300 & 300 & 300 \\ \hline
 Stock Prices & Simulated & Simulated & Real \\ \hline
 Correlation Matrix $(cm)$ & Random & Random & N/A \\ \hline
 No. Market Factors ($n_{exp}$) & 1 & 2 & N/A \\ \hline
 No. Experts & 8 & 8 & 8 \\ \hline
 $\mathrm{top}_k$ & 4 & 4 & 4 \\ \hline
\end{tabular}
\label{table:scparam}
\end{table}

\subsubsection{Market Scenario 1}
\label{subsubsec:scenario1}
Scenario 1 uses a simple setting where each market phase is governed by one \textit{single} market factor. We create 15 market phases of stock price data and divide them into 5 market regimes. Thus, each market regime comprises 3 market phases. Note that the 3 market phases within the same regime are generated from the same market factor (i.e., $(cm_i, exp_i ,vol_i)$), while the simulated price data in each phase are different due to the stochasticity of {MGBM}. Once the stock price data are generated, we create the historical {ADM} sequence $\mathcal{M}_t^{k,u}$ (Section \ref{subsec:adm_model}) as input to {ADNN} for forecasting assets dependencies. 

Figure \ref{fig:moe_visual}(a) depicts how experts are selected across the 15 market phases (market phases with the same regimes are posited adjacently, e.g. phases 1 to 3 correspond to the first regime, phases 4 to 6 the second regime, etc.). We can observe that the grid colors have similar distribution for market phases with the same market regime but different color distributions for phases within different regimes. This means that {MoE} picks similar combinations of experts for market phases with the same regime (sharing the same market factor), and different ones for market phases with different regimes, which aligns with the intuition.

\subsubsection{Market Scenario 2}
The real-world financial market is highly dynamic and is always jointly influenced by multiple factors. Scenario 1's single market factor assumption is too simple to simulate the real-world financial market. In scenario 2, we move one step forward and use two independent market factors to generate the stock price data for a better simulation of the complexity of the real market. 
From Figure \ref{fig:moe_visual}(a), we observe that experts selected in the second and third market regimes (corresponding to $\mbox{MP}^4\mbox{MP}^5\mbox{MP}^6$, and $\mbox{MP}^7\mbox{MP}^8\mbox{MP}^9$) of Scenario 1 tend to be mutually exclusive, indicating the low correlation of the two market factors in the two market regimes. 
Thus, in Scenario 2, we choose these two market regimes to generate the stock price data and {ADM}s.
Specifically, we utilize the {MGBM} theory to derive two log-return time series, $lr_2$ and $lr_3$, based on $(cm_2, exp_2 ,vol_2)$ and $(cm_3, exp_3 ,vol_3)$. Subsequently, we linearly combined these time series to generate {ADM}s in ten distinct market phases, as illustrated below:
\begin{equation}
\begin{aligned}
    \{\mbox{MP}^j = (1-w)\cdot lr_2 + w\cdot lr_3 \vert~& 0<j \leq 10, j \in \mathbb{N}, w = (j-1)/9\}
    \label{equ:sc2}
\end{aligned}
\end{equation}
Different $w$ indicate varying strengths of the two sub-factors, with $\mbox{MP}^1$ and $\mbox{MP}^{10}$ referring to the second and the third market regime, respectively.

Figure \ref{fig:moe_visual}(b) visualizes the results. 
We can see that in market phase $\mbox{MP}^1$ ($w=0$), the expert selection is similar to the second market regime in Scenario 1, where experts 1,2,3,5 are most frequently selected. 
In market phase $\mbox{MP}^{10}$ ($w=1$), the selection is closer to the third regime, where experts 4,5,7,8 are most frequently selected. 
As $w$ shifts from 0 to 1, we can see that the selected experts gradually transit from 1,2,3,5 to 4,5,7,8. 
This result demonstrates {ADNN}'s ability to detect and differentiate multiple market factors embedded in the input data and the capability of {MoE} to assign appropriate experts to model the market signals encoded in the corresponding market factors.

\subsubsection{Market Scenario 3}
In Scenario 3, we directly use the real-world price dataset as input to {ADNN} instead of using simulation data generated by {MGBM}.
We randomly sample 32 stocks from the 15-year US market data pool described in Section \ref{subsec:datasets} and divide the data into 15 market phases by year. 
Figure \ref{fig:moe_visual}(c) illustrates the result. Compared to Figures \ref{fig:moe_visual}(a) and \ref{fig:moe_visual}(b), we can observe that the expert distribution is much more diverse. This is reasonable since Scenario 3 includes all the market signals in the input data which are sophisticated and highly dynamic, requiring a wide range of experts with different knowledge to work collaboratively. 

\subsection{Visualization of the transformed {ADM}s}
\begin{figure}
  \centering
  \subfloat[The original {ADM}.]{\includegraphics[width=0.95\linewidth]{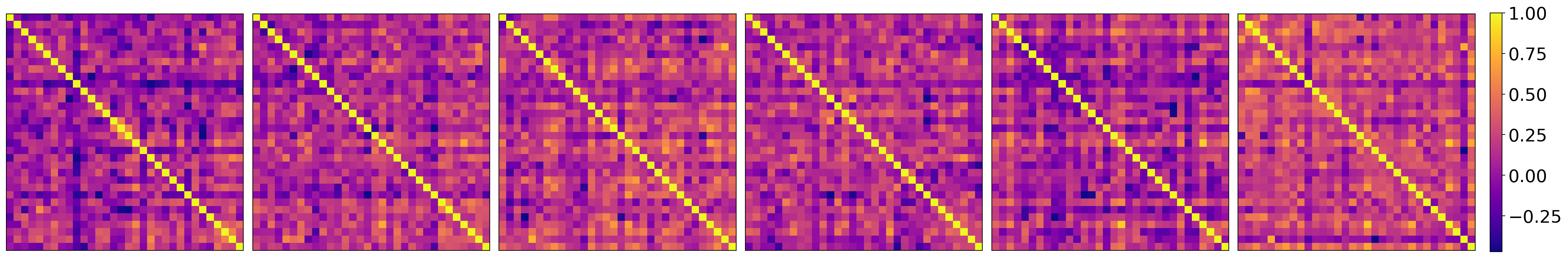}}\\
  \subfloat[The {ADM} after positional rearrangement.]
  {\includegraphics[width=0.95\linewidth]{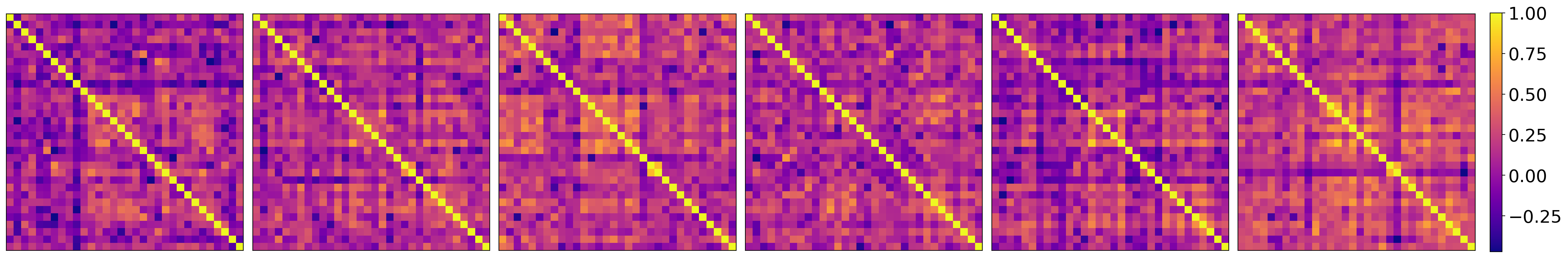}}\\
   \subfloat[The {ADM} after {MoE} transformation.]
  {\includegraphics[width=0.95\linewidth]{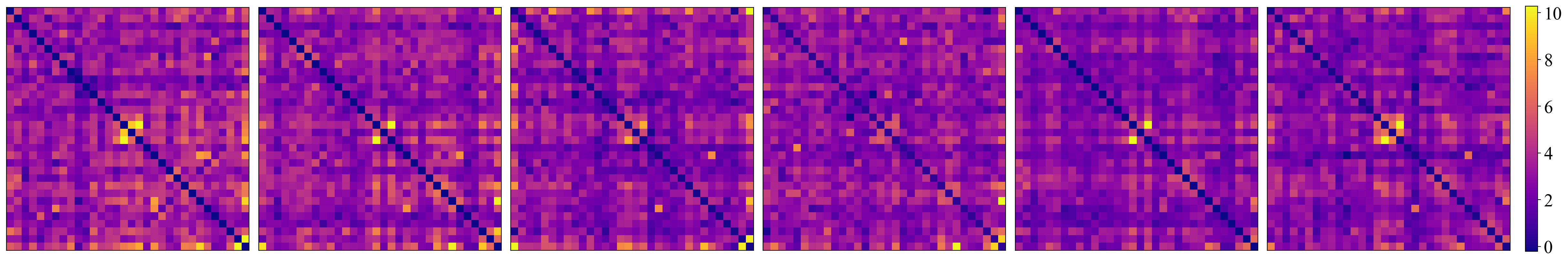}}
  \caption{{ADM} of 32 assets (dataset 1) in different periods.}
  \label{fig:visual_adm}
\end{figure}

We visualize the transformed {ADM}s and study how the positional rearrangement and the {MoE} transformation block improve the downstream {ADM} prediction task. 
Specifically, we train the \textit{PT-ADNN} model using dataset 1 and select six {ADM}s from the sequential training set, starting from index 0 of the {ADM} sequence with a uniform gap of size 100 between each consecutive {ADM}.
Figure \ref{fig:visual_adm} demonstrates the six {ADM}s in different periods.
We can observe that the positional rearrangement technique rearranges the row and column positions in such a way that assets exhibiting concurrent patterns are brought closer to each other. Take the third {ADM} image in Figure \ref{fig:visual_adm}(b) as an example. The resulting matrix bears a stronger resemblance to an image, featuring prominent, well-defined blocks with clear boundaries, each composed of uniform colors.
On the other hand, the {ADM}s transformed by the {MoE} method exhibit a wider range of values and higher contrast. It shows that the {MoE} transformation scales up the input values and introduces more divergence to the original data, which helps the downstream convolution layer better distill the information.

\subsection{Application Study}
In this section, we demonstrate how the predicted {ADM} can be used to solve two real-world financial problems.

\subsubsection{Portfolio Risk Reduction}
\label{subsec:app_portfoliio}
Portfolio diversification needs to consider the dynamic property of assets dependencies, based on which the weight of each asset in the portfolio can be obtained through the famous ``Modern Portfolio Theory'' \cite{Markowitz1952}. Equation \ref{equ:markov} briefly summarizes the idea:
\begin{equation}
\label{equ:markov}
\begin{aligned}
    \mathbf{\hat{\Sigma}}_m = &~\mathrm{diag} (\mathbf{s})\cdot \hat{\Gamma}_m \cdot \mathrm{diag}(\mathbf{s}) \\
    \mathbf{\omega}_{m} = &~\mathop{\arg\min}_{\omega} \mathbf{\omega}^T\hat{\mathbf{\Sigma}}_m\mathbf{\omega} \\
     \sigma_{m}^2 = &~ \mathbf{\omega}_{m}^T\mathbf{\Sigma}\mathbf{\omega}_{m}
\end{aligned}
\end{equation}

\noindent where $\hat{\Gamma}_m$ denotes the predicted correlation matrix for forecasting method $m$ (here, $\hat{\Gamma}_m=\hat{\mathcal{M}}_{t+h}$), $\textbf{s}$ denotes the ground truth standard deviation vector\footnote{Since the main purpose of this evaluation is to examine the performance of {ADM} (``correlation'') prediction, we assume prior knowledge of the future standard deviation vector to eliminate the extra variability.} at the time ($t+h$), diag($\textbf{s}$) is a diagonal matrix, $\hat{\mathbf{\Sigma}}$ denotes the predicted covariance matrix. 
We first calculate the predicted covariance matrix $\hat{\mathbf{\Sigma}}$ (first line of Equation \ref{equ:markov}) and obtain the asset weight vector $\omega_{m}$ (second line). Then, the actual variance of the derived portfolio $\mathbf{\omega}$ can be computed with the future ground-truth covariance matrix $\mathbf{\Sigma}$ (third line). 
In Equation \ref{equ:markov}, portfolio volatility $\sigma_{m}^2$ is a function of $\hat{\mathbf{\Sigma}}_m$, and when $\hat{\mathbf{\Sigma}}_m = \mathbf{\Sigma} $ the \textit{optimal} portfolio risk is obtained.

We randomly choose one stock dataset (dataset 7) as the source for constructing the diversified portfolio based on the predicted {ADM}s. All of the methods under evaluation build up their portfolios from the same $32$ stocks in dataset 7, which are then evaluated based on their respective portfolios' volatility $\sigma_{m}^2$. 
For baselines, we choose (1) the intuitive \textit{Previous} {ADM} (introduced at the beginning of Section \ref{subsec:eval_accuracy}), and (2) the well-known statistical method \textit{DCC-Garch}. 
We use \textit{T-ADNN} as it achieves the best performance in dataset 7. 
All baselines and the \textit{ADNN} will be compared to \textit{Optimal} ($\hat{\mathbf{\Sigma}} = \Sigma$). For simplicity, we assume that no short selling is allowed, i.e. asset weight $\omega_i \in [0,1]$.

\begin{figure}[tb]
  \centering
  \includegraphics[width=0.95\linewidth]{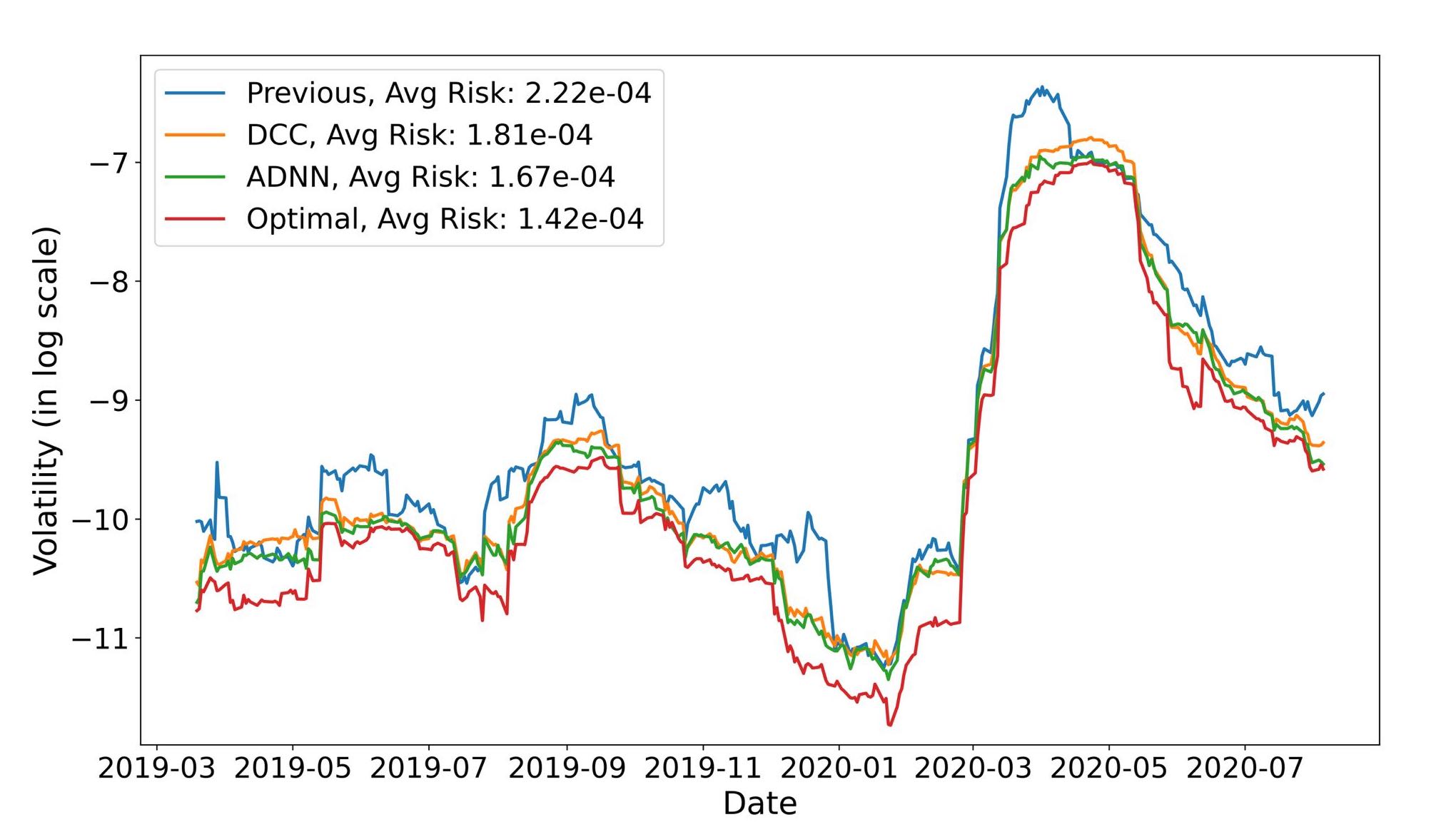} 
  \caption{Portfolio Example using Dataset 7.} 
  \label{fig:portfolio_example} 
\end{figure}

The comparison results are shown in Figure \ref{fig:portfolio_example}. We can observe that, as expected, both \textit{DCC} and \textit{ADNN} are superior to the method \textit{Previous}, and they display similar trends. \textit{ADNN} outperforms \textit{DCC} with an average risk of $1.67 \times 10^{-4}$, which ranks only second to the optimal portfolio. In terms of average risk, \textit{ADNN} achieves an improvement of $24.8\%$ over \textit{Previous} and $13.9\%$ over \textit{DCC}. In conclusion, the \textit{ADNN} method can generate a more diversified and lower-risk portfolio compared with the baselines.

\subsubsection{Pair Trading}
Pair trading is a trading strategy that involves the simultaneous buying and selling of two related securities, typically stocks, to take advantage of perceived price imbalances between them. 
Correlation plays a crucial role in pair trading strategies as it helps traders identify suitable pairs of assets to trade. A high correlation between two assets suggests that their prices move together, increasing the likelihood of finding profitable trading opportunities when the correlation temporarily breaks down. 
In this study, we utilize correlations predicted by \textit{ADNN} to inform decision-making in pair trading. Algorithm \ref{algo:pair_trading} illustrates the strategy.

\begin{algorithm}[t]
\caption{Correlation-based Pair Trading Strategy}\label{algo:pair_trading}
\begin{algorithmic}[1]
\renewcommand{\algorithmicrequire}{\textbf{Input:}}
\REQUIRE Two assets $a_{\alpha}$ and $a_{\beta}$, initial capital $C$, correlation threshold $\theta_c=0.7$, number of time periods $T$
\STATE Initialize $t=1$
\WHILE{$t < T$}
\STATE Predict the correlation coefficient $\hat{f}(a_{t+1, \alpha}, a_{t+1, \beta})$ for next time period $(t+1)$
\IF{$\hat{f}(a_{t+1, \alpha}, a_{t+1, \beta}) > \theta_c$}
\STATE Calculate the average price spread of the past year $s_p = \frac{1}{252}\sum_{i=1}^{252}(p_{t - i, \alpha} - p_{t-i, \beta})$
\IF{$(p_{t, \alpha} - p_{t, \beta})>s_p$  }
\STATE Allocate 50\% of capital to a long position in $a_{\alpha}$
\STATE Allocate 50\% of capital to a short position in $a_{\beta}$
\ELSE
\STATE Allocate 50\% of capital to a long position in $a_{\beta}$
\STATE Allocate 50\% of capital to a short position in $a_{\alpha}$
\ENDIF
\ELSE
\STATE Close the long/short position
\ENDIF
\STATE Update $t = t + 1$
\ENDWHILE
\end{algorithmic}
\end{algorithm}

The experiment is conducted using the randomly picked dataset 7, where the top ten asset pairs with a historically high correlation are selected. 
We evaluate multiple approaches by calculating the average performance metrics for pair trading across ten pairs, including the Win/Loss Rate, Maximum Drawdown, and Sharpe Ratio. 
We use the following common strategies for pair trading as baselines.
\begin{itemize}
    \item \textbf{Mean Reversion Strategy} assumes that the prices of the two assets in a pair will revert to their mean relationship. When the spread between the assets widens beyond a certain threshold, it takes a long position in the underperforming asset and a short position in the outperforming asset. The positions are closed when the spread returns to its average.
    \item \textbf{Statistical Deviation Strategy} takes positions expecting the spread to revert to its normal range when the spread between the assets exceeds a certain number of standard deviations or reaches extreme values. The positions are closed when the spread returns to a typical level.
    \item \textbf{S\&P 500} is a widely recognized stock market index comprising the 500 largest publicly traded companies in the United States. It serves as a benchmark for evaluating the effectiveness of the pair trading strategy compared to passive investing in the market.
\end{itemize}

Table \ref{tab:pair_trading} displays the results. 
The \textit{ADNN} model demonstrates superior performance in terms of the Win/Loss Rate, with a Sharpe ratio close to 2 and a maximum drawdown below 20\%.  
This outcome indicates that the predicted correlation effectively captures market changes and validates the efficacy of the \textit{ADNN} when applied to pair trading.

\begin{table}[t]
\centering
\footnotesize
\caption{Performance Comparison on Pair Trading.}
\setlength{\tabcolsep}{3pt}
\begin{tabular}{c||c|c|c|c}
\hline
\multirow{2}*{\diagbox[]{Method}{Metrics}} &  Profit & Maximum & Sharpe&Win/Loss\\
&Rate&Drawdown&Ratio&Rate \\\hline
Mean Reversion& 7.3\%&  5.6\%&1.5&65\%\\\hline
Statistical Deviation& 5.3\%&  4.1\%&2.1&69\%\\\hline
S\&P 500&8.4\%&33.9\%&0.45&-\\\hline
ADNN& 12.1\%&  10.7\%&2.0&68\%\\\hline
\end{tabular}
\label{tab:pair_trading}
\end{table}

\section{Conclusion}\label{sec:conclusion}
Prediction of asset dependency has been extensively researched in the financial industry. In this paper, we formulate the problem as the sequential Asset Dependency Matrix (\textit{ADM}) prediction problem. We propose the Asset Dependency Neural Network (\textit{ADNN}) framework to model the spatiotemporal dependencies signals in the historical {ADM} sequence to predict future {ADM}. To resolve the {ADM} representation problem, {ADNN} incorporates a Mixture of Experts (\textit{MoE}) to transform input {ADM} to the optimal representation to facilitate {ConvLSTM} model in predicting the future {ADM}s. We validate and explain the effectiveness of {ADNN} on multiple real-world stock market data by comparing it with various baselines and applying it to the portfolio diversification task. 
For future work, we plan to extend the current research in the following three directions: (1) investigate other end-to-end transformation methods, (2) apply the {ADNN} framework to other kinds of {ADM} (e.g., covariance matrix), and (3) construct a portfolio with the consideration of realistic constraint (e.g., transaction cost).  

\section*{Acknowledgements}
The research reported in this paper was supported by the Research Grants Council HKSAR GRF (No. 16215019), Guangdong Provincial Key Laboratory of Interdisciplinary Research and Application for Data Science,
BNU-HKBU United International College, project code
2022B1212010006, and in part by Guangdong Higher Education Upgrading Plan (2021-2025) of ``Rushing to the Top, Making Up Shortcomings and Strengthening Special Features'' with UIC research grant R0400001-22.

\section*{Declaration of Generative AI and AI-assisted technologies in the writing process}
During the preparation of this work, the authors used ChatGPT to improve the readability and language of the manuscript. After using this service, the authors reviewed and edited the content as needed and take full responsibility for the content of the publication.



\bibliographystyle{elsarticle-harv} 
\bibliography{citation}






\end{document}